\documentclass[pdflatex,sn-mathphys-num]{sn-jnl}

\usepackage{graphicx}%
\usepackage{multirow}%
\usepackage{amsmath,amssymb,amsfonts}%
\usepackage{amsthm}%
\usepackage{mathrsfs}%
\usepackage[title]{appendix}%
\usepackage{xcolor}%
\usepackage{textcomp}%
\usepackage{manyfoot}%
\usepackage{booktabs}%
\usepackage{algorithm}%
\usepackage{algorithmicx}%
\usepackage{algpseudocode}%
\usepackage{listings}%
\usepackage{makecell}
\usepackage{cleveref}
\usepackage{placeins}

\begin{document}

\title[Scalable, Trustworthy and Energy-Efficient Predictive Graph Foundation Models]{Scalable Training of Trustworthy and Energy-Efficient Predictive Graph Foundation Models for Atomistic Materials Modeling: \\ A Case Study with HydraGNN}


\author*[1]{\fnm{Massimiliano} \sur{Lupo Pasini}}\email{lupopasinim@ornl.gov}

\author[2]{\fnm{Jong Youl} \sur{Choi}}\email{choij@ornl.gov}

\author[3]{\fnm{Kshitij} \sur{Mehta}}\email{mehtakv@ornl.gov}

\author[4]{\fnm{Pei} \sur{Zhang}}\email{zhangp1@ornl.gov}

\author[5]{\fnm{David} \sur{Rogers}}\email{rogersdm@ornl.gov}

\author[6]{\fnm{Jonghyun} \sur{Bae}}\email{jbae2@lbl.gov}

\author[7]{\fnm{Khaled Z.} \sur{Ibrahim}}\email{kzibrahim@lbl.gov}

\author[8]{\fnm{Ashwin M.} \sur{Aji}}\email{ashwin.aji@amd.com}

\author[9]{\fnm{Karl W.} \sur{Schulz}}\email{karl.schulz@amd.com}

\author[10]{\fnm{Jord\`{a}} \sur{Polo}}\email{jorda.polo@amd.com}

\author[11]{\fnm{Prasanna} \sur{Balaprakash}}\email{pbalapra@ornl.gov}

\affil[1, 4]{\orgdiv{Computational Sciences and Engineering Division}, \orgname{Oak Ridge National Laboratory}, \orgaddress{\street{1 Bethel Valley Road}, \city{Oak Ridge}, \postcode{37831}, \state{TN}, \country{USA}}}

\affil[2, 3]{\orgdiv{Computer Science and Mathematics Division}, \orgname{Oak Ridge National Laboratory}, \orgaddress{\street{1 Bethel Valley Road}, \city{Oak Ridge}, \postcode{37831}, \state{TN}, \country{USA}}}

\affil[5]{\orgdiv{National Center for Computational Sciences}, \orgname{Oak Ridge National Laboratory}, \orgaddress{\street{1 Bethel Valley Road}, \city{Oak Ridge}, \postcode{37831}, \state{TN}, \country{USA}}}

\affil[6, 7]{\orgdiv{Computer Science Department}, \orgname{Lawrence Berkeley National Laboratory}, \orgaddress{\street{1 Cyclotron Road}, \city{Berkeley}, \postcode{94720}, \state{CA}, \country{USA}}}

\affil[8, 9, 10]{\orgdiv{Advanced Micro Devices}, \orgname{AMD Research}, \country{USA}}

\affil[11]{\orgdiv{Computing and Computational Sciences Directorate}, \orgname{Oak Ridge National Laboratory}, \orgaddress{\street{1 Bethel Valley Road}, \city{Oak Ridge}, \postcode{37831}, \state{TN}, \country{USA}}}


\abstract{We present our work on developing and training scalable, trustworthy, and energy-efficient predictive graph foundation models (GFMs) using HydraGNN, a multi-headed graph convolutional neural network architecture. HydraGNN expands the boundaries of graph neural network (GNN) computations in both training scale and data diversity. It abstracts over message passing algorithms, allowing both {\em reproduction of} and {\em comparison across} algorithmic innovations that define nearest-neighbor convolution in GNNs. This work discusses a series of optimizations that have allowed scaling up the GFMs training to tens of thousands of GPUs on datasets consisting of hundreds of millions of graphs. Our GFMs use multi-task learning (MTL) to simultaneously learn graph-level and node-level properties of atomistic structures, such as energy and atomic forces. Using over 154 million atomistic structures for training, we illustrate the performance of our approach along with the lessons learned on two state-of-the-art United States Department of Energy (US-DOE) supercomputers, namely the Perlmutter petascale system at the National Energy Research Scientific Computing Center and the Frontier exascale system at Oak Ridge Leadership Computing Facility. The HydraGNN architecture enables the GFM to achieve near-linear strong scaling performance using more than 2,000 GPUs on Perlmutter and 16,000 GPUs on Frontier. 
}




\maketitle

{\footnotesize \noindent This manuscript has been authored in part by UT-Battelle, LLC, under contract DE-AC05-00OR22725 with the US Department of Energy (DOE). The US government retains and the publisher, by accepting the article for publication, acknowledges that the US government retains a nonexclusive, paid-up, irrevocable, worldwide license to publish or reproduce the published form of this manuscript, or allow others to do so, for US government purposes. DOE will provide public access to these results of federally sponsored research in accordance with the DOE Public Access Plan (\url{http://energy.gov/downloads/doe-public-access-plan}).}

\section{Introduction}

Discovery of new materials with desired properties, accurate predictions of a material's behavior throughout its entire lifespan is crucial to fundamental scientific progress
in energy generation, transportation, electronics, and information technology \cite{clean_energy_doe_newsletter}. 
Machine learning (ML) has shown great potential in accelerating the screening and pre-selection of materials for further experimental testing. In particular, deep learning (DL) models effectively captured relevant underlying relationships due to the arrangement of atoms of different constituents within various atomistic structures \cite{Balabin, Chandrasekaran, Brockherde, Sinitskiy, custodio, schleder, kalinin2019, maguire2001, Wang, pasini, park2021accurate}. DL models trained on data generated from experiments and/or first-principles calculations can predict the properties of interest to be used as new inputs. This inference takes only a fraction of the time it would take to run an experiment or a full first-principles calculation while still producing sufficiently accurate results. This drastic reduction in time to predict material properties using atomistic information results in a promising path towards accelerating material discovery and design \cite{axelrod2022learning, yoo2023deep}.

However, generating vast volumes of experimental and/or first-principles training data is impractical even with sophisticated experimental facilities and powerful supercomputers. 
Recently, foundation models (FMs) have shown the promise to navigate around this challenge: once pre-trained over a large volume of available open-source data \cite{zhang2024scientific}, an FM would provide a jump-start to refined models by fine-tuning on significantly smaller amounts of data for customized applications (also called downstream tasks). 
Reducing the number of simulations and/or experiments for generating domain-specific training data also drastically reduces the energy costs of developing domain-specific DL models. 



While state-of-the-art language-based FMs with a transformer architecture have reached promising results in several domains \cite{vaswani2017attention, devlin2019bert, liu2019roberta, raffel2020exploring, brown2020language, webersinke2021climatebert, tshitoyan2019matbert, beltagy2019scibert, kim2020polymerbert, chithrananda2020chemberta, scibert2023, Kuenneth2023polyBERT}, they fail to 
capture important topological aspects of the atomistic structures. Therefore, alternative DL architectures need to be considered for the development of trustworthy (interpreted as simultaneously accurate and highly confident) FMs for materials using atomistic scale information.  

Since atomistic material structures for a generic type of compound can be mapped onto a graph (where atoms can be treated as nodes and interatomic bonds as edges), graph foundation models (GFMs) \cite{mao2024positiongraphfoundationmodels, shi2024}, which are FMs that operate on data structures as graphs, are the candidate of choice for these applications. 
Currently, GFMs proposed in the literature are developed by training graph neural network (GNN) architectures on a sufficiently large and comprehensive dataset for the domain of interest. 
While a few efforts have already been undertaken to develop GFMs for atomistic materials modeling applications \cite{takeda2023multimodal, 10.1145/3624062.3626081, beaini2024towards, Sypetkowski2024}, the existing work is still at an incipient stage.
Current efforts do not yet ensure that their proposed approach achieves trustworthiness.
Moreover, serious concerns have been recently raised about the unaffordable amount of energy requested by proliferating AI centers due to computationally intensive tasks of training large FMs on large volumes of data. 

The work described in this manuscript developed scalable and trustworthy supervised GFMs for the simultaneous prediction of energies and atomic forces with careful attention to energy consumption.
The GFMs have been constructed using HydraGNN \cite{hydragnn3, usermanual_hydragnn}, a fully scalable GNN architecture developed at ORNL. 
Experiments were conducted on two large US-DOE supercomputers: the Perlmutter petascale machine at National Energy Research Scientific Computing Center (NERSC) and the Frontier exascale system at Oak Ridge Leadership Computing Facility (OLCF).

The remaining content of the paper is organized as follows.
We discuss the current state of the art and introduce HydraGNN in Section \ref{sec:sota}. In Section \ref{sec:main-contribution}, we discuss our approach toward developing a scalable framework and list the different optimizations for scalable training. We discuss our use of large-scale HPO to develop a trustworthy GFM. Section \ref{performance_section} shows the performance of different components of this work: reading large data, scaling the training process, performing HPO at large scale, and measuring epistemic uncertainty using an ensemble of HPO trials with high predictive accuracy and low energy costs. We conclude our study and discuss future work in Section \ref{sec:conclusion}.

\section{Current state-of-the-art}
\label{sec:sota}

\subsection{GNN training on open-source atomistic materials data}
To date, there have been a few approaches proposed in the literature to develop GFMs for atomistic materials modeling. In \cite{takeda2023multimodal}, the authors proposed a multi-modal approach where multiple encoding DL architectures are trained on different types of data representations and describing different physical quantities. The models are aligned to each other through a penalization term in the training loss function that forces latent vectors from each embedding space to coincide. However, the datasets used comprise only organic molecules, which cover only a relatively small set of natural elements on the periodic table. 

In \cite{beaini2024towards}, the authors collected open-source datasets that provide labels for different properties of organic molecules. Using such a diverse collection of datasets, a GNN architecture is used for MTL in order to identify embedding spaces that capture meaningful correlations between the different labeled properties, with the promise that such an embedding space would reduce the data requirement on downstream tasks specific to organic chemistry. Since the model is trained on open-source datasets that describe only organic molecules, this approach is not transferable to inorganic compounds. Moreover, the authors compare the performance of different message passing neural network (MPNN) layers to construct the GNN architecture by performing computationally inexpensive hyperparameter tuning on small models with few parameters and transfer the use of such hyperparameters to models of much larger scale. While this approach helps limit the computational burden of HPO on large scale GFMs, the best performing configuration of hyperparameters at small scale is not guaranteed to be the best performing configuration of hyperparameters at a larger scale and on a larger set of data, because the conclusions drawn from the HPO study are model and data dependent. 

In \cite{batatia2024foundation}, the authors developed a GFM trained on the Materials Project Trajectories (\texttt{MPTrj}) dataset \cite{mptrj}, using an MPNN layer that is capable of modeling 4-body interactions. As the authors themselves recognize in their conclusions, while the approach sheds light onto a promising path towards building effective GFMs for atomistic materials modeling, the impact of their work is limited by the fact that the GFM has a very small number parameters that was deliberately maintained low due to computational limitations. Moreover, this reduces the expressivity of the GFM. 

In \cite{NEURIPS2023_5f02c76b}, the authors propose a new kernel function for GNNs for modeling potential energy surfaces,
which exploits chemical species information, and illustrate its efficacy in the context of transferable GFMs by pre-training the GFM on the Open Catalyst 2020 (\texttt{OC2020}) dataset \cite{oc2020} and illustrating its performance on a set of fine-tuning tasks. Since the GFM was pre-trained only on the \texttt{OC2020} dataset, the applicability of this GFM is restricted to inorganic compounds. 

While not explicitly presented by their developers as GFMs, there have been other models that cover broader sets of elements of the periodic table compared to the approaches mentioned in the previous paragraphs. 
In \cite{10.1145/3624062.3626081}, the authors built a GNN model using MTL for simultaneous predictions of several material properties by training the GNN model on multiple datasets, including \texttt{OC2020} \cite{oc2020} and Open Catalyst 2022 (\texttt{OC2022}) \cite{oc2022}. However, the approach considers only a single GNN architecture without performing HPO. Moreover, the set of parameters in the GNN model is relatively small, of the order of few millions of parameters, which limits the attainable accuracy on large volumes of data. 

In \cite{Sypetkowski2024}, the authors studied the scaling behavior of 2D molecular GNNs under varied settings of depth, width, number of molecules, number of labels, the diversity in dataset,
and architectural choice. The authors showed that supervised pre-training of large GNNs on molecular datasets provides a rich fingerprint embedding, which is useful for 38 downstream tasks. Even if this work very systematically studied the effect of GNN model size over the predictive performance in the pre-training and fine-tuning stage with many and diverse downstream tasks, the work has two important limitations: it only considers 2D graphs and it addressed only organic compounds. 

Several UQ methods have been applied to GNNs \cite{hirschfeld2020uncertainty}, including Bayesian GNNs \cite{ryu2019bayesian}, prediction interval methods \cite{huang2024uncertainty}, and deep ensemble methods \cite{jiang2023uncertainty}.
Bayesian methods are theoretically rigorous but challenging to scale to high-dimensional data. 
Prediction interval methods are cost-effective but often require tedious tuning of heuristic parameters. 

Compared to the scientific contributions mentioned above, our work distinguishes itself by leveraging extreme scale supercomputing resources to ensure trustworthiness of the GFMs by performing (i) a systematic large scale HPO across a broad set of GNN architectures and (ii) a large scale ensemble learning (EL) for UQ, which realizes a compromise between cost and performance.

\subsection{Scalability and GPU optimization for GNN training}
The effect of the specific algorithmic characteristics of GNNs on performance benchmarking has been carried out on GPUs \cite{9408205}, where the authors noted that GNN training differs significantly from conventional
convolutional networks (CNNs) in that only 25\% of the execution
time is spent on dense and sparse matrix multiplications compared
to 50\% in CNNs. Moreover, the execution time to process graph samples in GNNs was noted to vary greatly according to the size of the graph (number of nodes and number of edges) of the input data.
The studies conducted in this work showed that the majority of the time during GNN training was spent in integer operations, sorting, index selection, reductions, and scatter-gather operations needed for nodal and edge feature updates with message passing.
Multi-GPU scaling was reported using up to 4 GPUs, showing
about 20-50\% strong scaling efficiency between 1 and 4 GPUs.
Similar remarks apply to refs.~\cite{10.1145/3419111.3421281, 10046129, balin2022mggcn, md2021distgnn}, which characterize
subdivision of large graphs among processors and parallel aggregation
during convolution steps. 

These are useful conclusions for optimization of GNN training on large graphs (i.e., with millions of nodes), but need to be re-evaluated for our datasets.
Training on large graphs can be highly sensitive to the splitting scheme
used to partition the graph into subgraphs and to distribute them among processors.
For the atomistic materials modeling applications addressed in our work,
the graph samples are small (with at most a few hundreds of nodes).
For GNN convolutions in particular, convolution on a batch
of samples will have a much more local, block diagonal structure.
Therefore, throughput should be less sensitive to the choice of atomistic structures per batch.

Using a larger number of GPUs, the developers of the PyTorch framework for distributed data parallelism (DDP) showed the benefit of overlapping computation with communication, showing near-linear scaling using up to 256 NVIDIA Tesla V100 GPUs \cite{li2020pytorch}.
These preliminary scaling results focused on DDP for training of DL model using a moderate volume of data.




\subsection{HydraGNN}
The complexity of the physics and the scale at which atomistic structures must be studied in response to US-DOE needs in materials science makes it compelling to develop GNN capabilities that simultaneously satisfy several important algorithmic and computer science requirements. To effectively respond to the scientific needs of the US-DOE, a GNN architecture must provide: (1) capabilities to read and process data from multiple sources simultaneously, (2) flexibility to support diverse DOE-relevant scientific applications, (3) capabilities to scale the training on leadership class supercomputing facilities, (4) portability across heterogeneous computing environments, (5) continuous software maintenance by ensuring support and compatibility with upgraded software dependencies, and (6) maintained documentation to support new users across a broad set of international institutions.  

While several GNN architectures have been made available as open-source tools to the scientific community in the last few years \cite{viniavskyi2022openglue, openmatscimltoolkit, lee2023matsciml, doi:10.1021/acsomega.1c04017}, none of these tools satisfies all of the above requirements. Moreover, including missing capabilities on these well-established GNN libraries requires invasive and laborious modifications for software re-design.
These challenges arising from existing GNN implementations motivated our effort in developing HydraGNN \cite{hydragnn3, usermanual_hydragnn}. In response to the US-DOE scientific needs, HydraGNN provides:
\begin{itemize}
\item MTL capabilities to process multi-source, multi-fidelity data \cite{Lupo_Pasini_2022};
\item object-oriented programming capabilities to use different MPNN layers \cite{10500060}, which allows flexible switching between different message policies based on the scientific needs of the specific application at hand, as well treating the MPNN layer as a tunable categorical hyperparameter with HPO;
\item invariant and equivariant features that reduce computational redundancy and time-to-solution \cite{baker2023invariant}, therefore contributing to energy saving;
\item scalable input/output (I/O) data management techniques to efficiently scale the training of GNN models on millions of data samples using thousands of GPUs on supercomputing facilities \cite{choi2022scalable}; and
\item portable capabilities that allow conveniently running the GNN training on diverse computing platforms with different hardware and software specifications.
\end{itemize}


HydraGNN uses the PyTorch~\cite{pytorch, pytorch2019} software for automatic differentiation and the PyTorch Geometric~\cite{fey_2019, torch_geometric} software for message passing. 
The architectural hyperparameters that determine the HydraGNN model size and complexity can be set in a configuration file to tune the model training and inference process easily.
Overall, HydraGNN is developed and maintained as a high-quality software product for large scale training and development of ML models \cite{hydragnn3}.

\section{Our contribution}
\label{sec:main-contribution}

The work described in this manuscript is the first-of-its-kind large-scale training of GFMs for atomistic materials
modeling using over 154 million atomistic structures as data samples and using over $91\%$ of the exascale supercomputer Frontier. We have used three key techniques for developing a scalable and trustworthy GFM: 1) scalable data management using a scientific data management library and an in-memory data store, 
2) scalable HPO that uses asynchronous Bayesian optimization for efficiently managing computing resources, and 
3) ensemble methods for UQ that allows model generalization and concurrently training multiple models. 
These three advancements collectively enhance the robustness, efficiency, and scalability of the GNN training process.


Compared to previous studies, our work shows near-linear scaling using 10x more GPUs and using much larger volumes of data, which introduces important challenges in I/O that we addressed to reduce computational bottlenecks and minimize communication overheads. Moreover, our results are generated using 
GPUs of newer generations, namely NVIDIA A100 installed on NERSC-Perlmutter and AMD Instinct\texttrademark{} MI250x installed on OLCF-Frontier, thereby showing that our scaling efficiency is also transferable across technologies manufactured by different vendors. 

\subsection{Data aggregation}

\begin{table}[ht!]
\normalsize
\centering
\begin{tabular}{|l|r|r|}
\hline
\textbf{Dataset} & \textbf{Number of data samples} & \textbf{Size} \\
\hline
\texttt{ANI1x} \cite{Smith2020ANI1ccx} & 4,956,005  & 5.3 GB\\
\texttt{QM7-X} \cite{qm7x} & 4,195,237 & 23 GB \\
\texttt{OC2020} \cite{oc2020} & 134,929,018 & 4.3 TB \\  
\texttt{OC2022} \cite{oc2022} & 8,847,031 & 648 GB \\
\texttt{MPTrj} \cite{mptrj} & 1,580,395 & 17 GB \\
\hline
Total & 154,507,686 & 5.2 TB \\
\hline
\end{tabular}
\vspace{0.1in}
\caption{Overview of Datasets used for training HydraGNN}
\label{tab:dataset}
\end{table}

Using large datasets for GFM training can enhance generalizability and ensure resilience to data variance issues that typically arise during downstream tasks.
To this end, we aggregated five open-source atomistic materials modeling datasets that are extremely diverse in terms of chemical composition, atomistic configurations, and number of atoms in the system. 
These datasets, as listed in Table~\ref{tab:dataset}, are: \texttt{ANI1x}, \texttt{QM7x}, \texttt{OC2020}, \texttt{OC2022}, and \texttt{MPTrj}.
\begin{itemize}
\item \texttt{ANI1x} \cite{Smith2020ANI1ccx} consists of over 4,956,005 atomistic structures derived from up to 57 thousand distinct molecular configurations containing the C, H, N, and O chemical elements.
\item \texttt{QM7x}  \cite{qm7x} is a comprehensive dataset of 42 physicochemical properties for approximately 4.2 million equilibrium and non-equilibrium structures of small organic molecules with up to seven non-hydrogen atoms from the C, N, O, S, Cl chemical elements.
\item \texttt{OC2020}  \cite{oc2020} provides 1,281,040 density functional theory (DFT) relaxations (134,890,000 single point calculations) across a range of oxide materials, coverages, and adsorbates. 
\item \texttt{OC2022}  \cite{oc2022} provides 62,331 DFT relaxations (9,854,504 single point calculations) across a range of oxide materials, coverages, and adsorbates.
\item \texttt{MPTrj} \cite{mptrj}: the version of the dataset from 2020 provides DFT calculations for 83,988 atomistic structures of inorganic materials. 
\end{itemize}

\begin{figure}[ht!]
    \centering
    \includegraphics[width=0.49\textwidth]{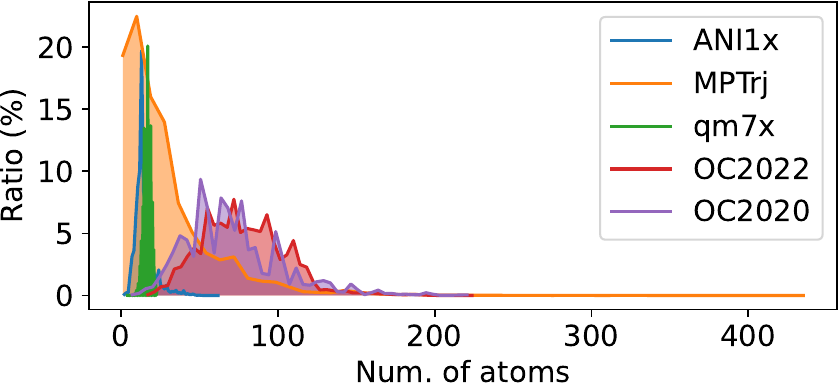}
    \includegraphics[width=0.48\textwidth]{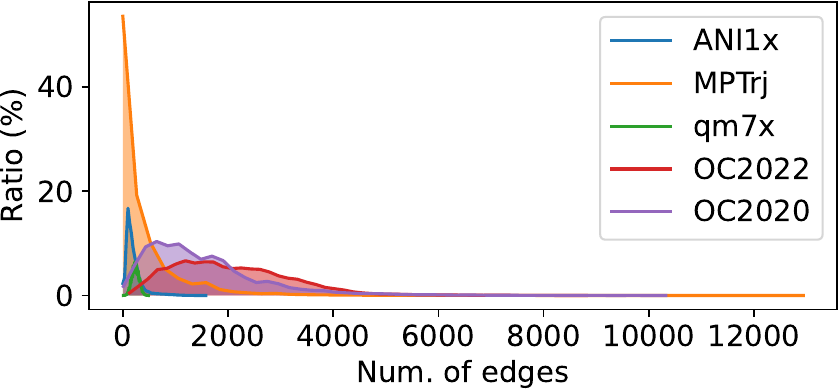}
    \caption{Normalized histograms for each dataset of the number of atoms within an atomistic structure (top) and number of edges within the graph representation of each atomistic structure (bottom). }
\label{fig:hist_atoms_edges}
\end{figure}

Each dataset is unique for the chemical compositions and the number of atoms in the atomistic structures of the compounds described. Figure~\ref{fig:hist_atoms_edges} shows the distribution of the number of atoms and graph edges per atomistic structure for each dataset.
For the \texttt{MPTrj} dataset, approximately half of the atomistic structures are relatively small in size. On the other hand, the \texttt{OC2020} and \texttt{OC2022} datasets consist of a more even distribution of atomistic structures with different sizes and edge counts, with the larger structures consisting of over 400 atoms and over 12,500 edges.
In total, the data used for training, validating, and testing our GFM consisted of over 154 million atomistic structures that consume 5.3 Terabytes of storage space.
These datasets were pre-processed using a scientific data management library into a common format for efficient storage and I/O, as discussed in Section \ref{adios-io}.

\begin{figure*}
    \centering
    \includegraphics[width=0.95\textwidth]{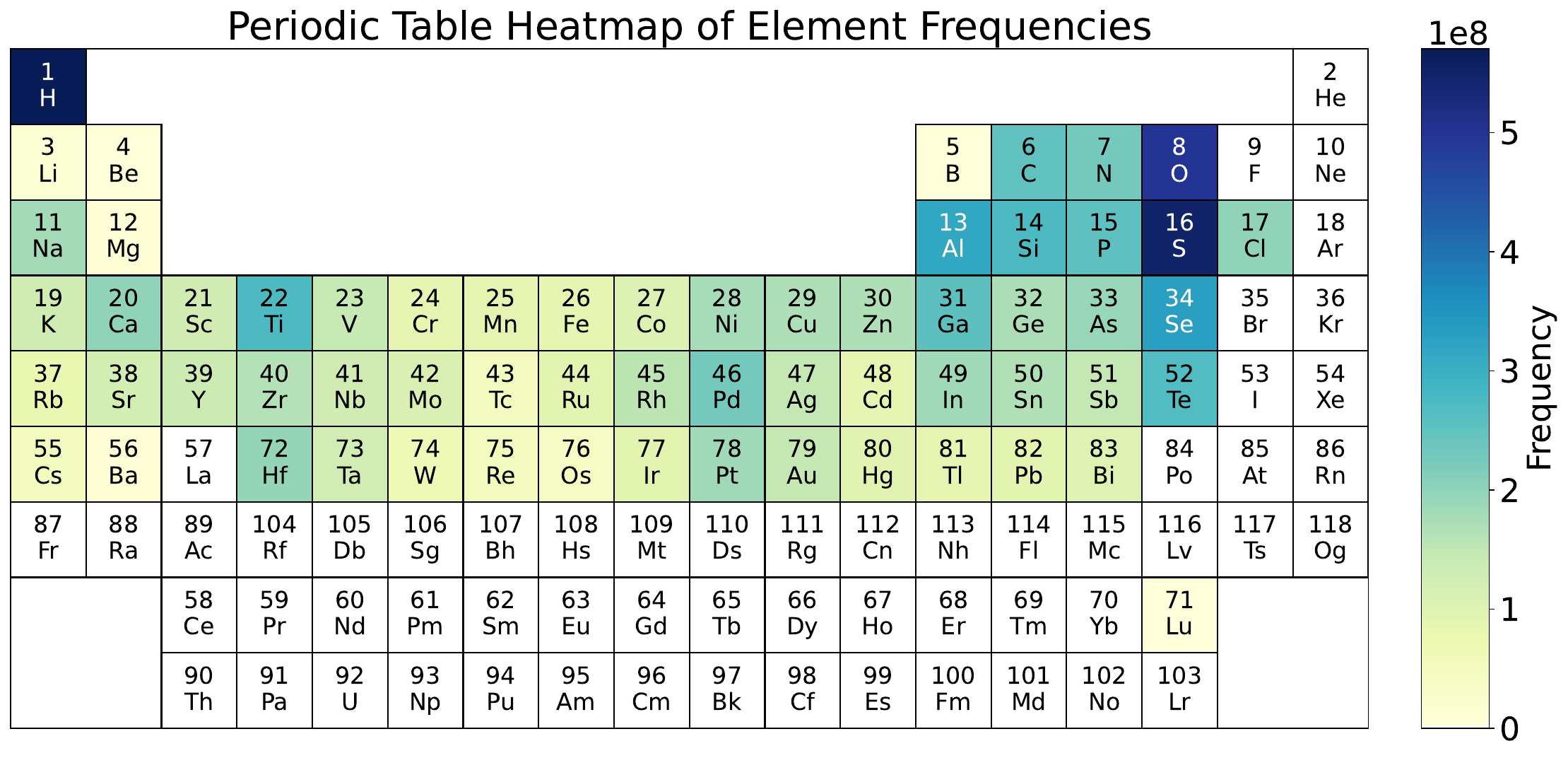}
    \caption{Heatmap that describes the frequency of occurrence of each element of the periodic table across data sampled resulting from the aggregation of the datasets ANI1x, QM7-X, OC2020, OC2022, and MPTrj.}
\label{fig:heatmap}
\end{figure*}

Figure \ref{fig:heatmap} provides the heatmap that illustrates the frequency of occurrence of each element of the periodic table across the entire dataset that results from the aggregation of the datasets \texttt{ANI1x}, \texttt{QM7-X}, \texttt{OC2020}, \texttt{OC2022}, and \texttt{MPTrj}. We notice that there is an under-representation of the elements from the transition metal groups. This is partially due to the fact that first-principles calculations for materials including these elements (e.g., alloys) are more computationally expensive, and thus are available in smaller volumes with respect to first-principles data for other classes of materials. 

\subsection{Data cleaning and pre-processing}
\label{data-cleaning}
Some of the atomistic structures were determined to have unrealistic values for atomic forces (on the order of 20,000 eV/angstrom).
These probably corresponded to configurations visited at early stages
of energy minimization during the generation of these datasets.
To eliminate these outliers, we first applied a data cleaning operation in which we discarded all atomistic structures with an $L2$-norm (also known as spectral norm) of the force tensor above 100 eV/angstrom to ensure that these data samples did not affect the training of our GFMs. 
The number of data samples removed from each dataset by this filtering operation is reported in Table \ref{tab:dataset_discarded}. 
\begin{table}[ht!]
\normalsize
\centering
\begin{tabular}{|l|r|}
\hline
\textbf{Dataset} & \textbf{Number of data samples removed} \\
\hline
\texttt{ANI1x} \cite{Smith2020ANI1ccx} & 0 \\
\texttt{QM7-X} \cite{qm7x} & 0 \\
\texttt{OC2020} \cite{oc2020} & 1 \\  
\texttt{OC2022} \cite{oc2022} & 12,270 \\
\texttt{MPTrj} \cite{mptrj} & 151 \\
\hline
Total & 12,422 \\
\hline
\end{tabular}
\vspace{0.1in}
\caption{Number of data samples discarded from each dataset due to values of the $L2$-norm (also known as spectral norm) of the force tensor being unreasonably over 100 eV/angstrom.}
\label{tab:dataset_discarded}
\end{table}

Datasets from different sources were generated with different electronic structure methods,
leading to global shifts in the energy calculated for each element of the periodic table, and consequently also on the energies of the organic and inorganic compounds obtained by combining multiple elements. 
In order to re-align the multi-source multi-fidelity data, we adopted the procedure proposed in \cite{batzner2024} to transform the energy of each atomistic structure during pre-processing by subtracting a linear regression term.
For each dataset, the linear regression term was calculated by solving the following least-squares problem:
\begin{equation}
\mathop{\mathrm{arg\,min}}_{C_1, \ldots, C_{118}} \sum_{i=1}^{N_{data}}\bigg ( e^i_0 - \sum_{Z=1}^{118} C_Z n^i_Z \bigg )^2
\end{equation}
where $e^i_0$ is the reference energy for the atomistic structure $i$, and $n^i_Z$ counts the number of atoms of element number $Z$ belonging to the atomistic structure $i$.
By separately determining the regression coefficients $C_Z$ (with the dimension unit of eV/atom) for each dataset, we were able to remove the largest source of variability between datasets. This led to a more consistent training of the GFMs using the labeled energy values.
Even if the datasets cover two-thirds of the natural elements of the periodic table, we built the linear regression model to account for all the 118 elements (natural and artificial) of the periodic table to ensure that our software infrastructures accommodates future generalizations of our current approach. 

\begin{figure}[ht!]
    \centering
    \includegraphics[width=0.48\textwidth]{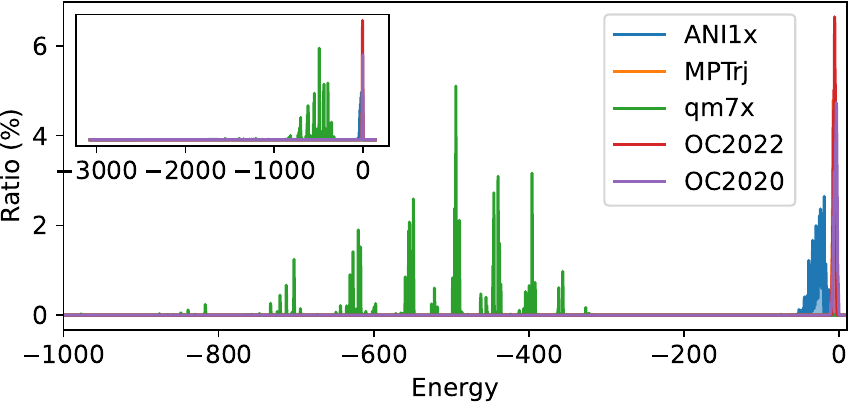}\\
    \vspace{0.05in}
    \includegraphics[width=0.48\textwidth]{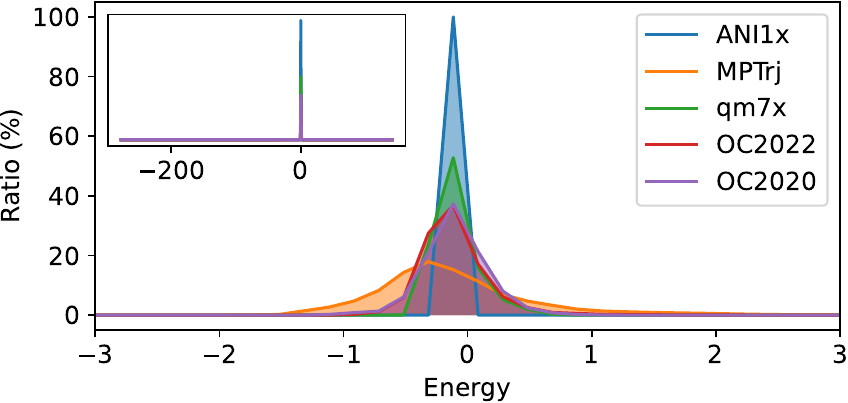}\\
    \vspace{0.05in}
    \includegraphics[width=0.48\textwidth]{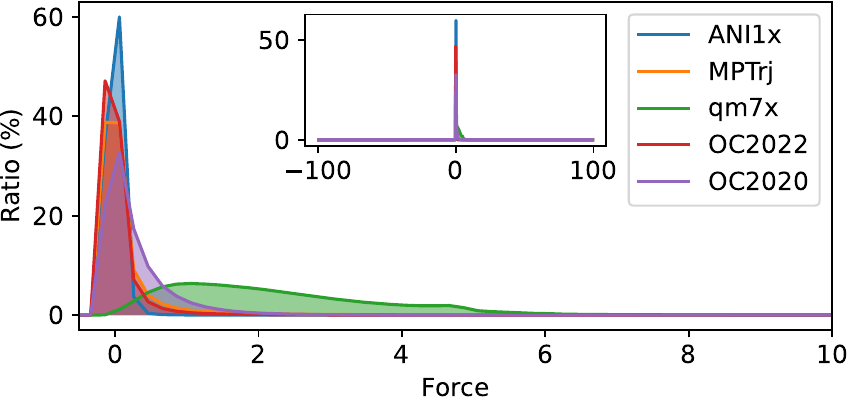}
    \caption{Normalized histograms for each dataset of the total energy per atom (top)
    , energy per atom after removing the linear regression term from each dataset (center), 
    and the $L^2$-norm of the force tensor (bottom) for each dataset after removing data samples with $L^2$-norm of the force tensor unreasonably higher than 100 eV/angstrom. }
\label{fig:hist_energy_forces}
\end{figure}

\begin{figure*}
    \centering
    \includegraphics[width=0.99\textwidth, trim=0pt 0pt 0pt 5pt, clip]{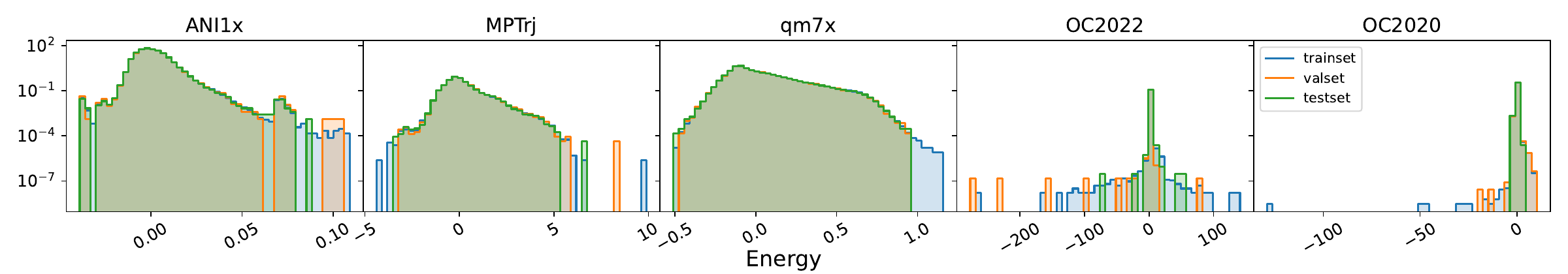}
    \includegraphics[width=0.99\textwidth, trim=0pt 3pt 0pt 25pt, clip]{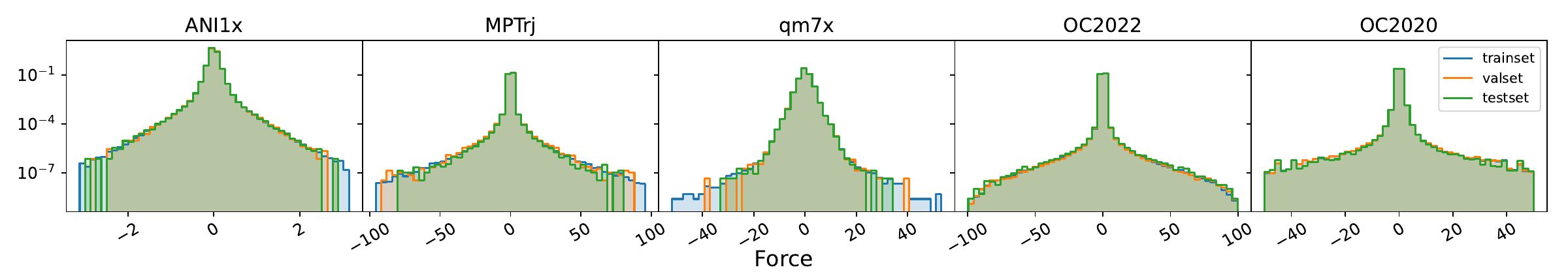}
    \caption{Histograms showing the energy and force distributions across five datasets, along with the distribution breakdown for the training, validation, and test sets.
    The y-axis shows the probability density on a logarithmic scale.}
\label{fig:hist_3set}
\end{figure*}

The histogram of the values of the energy per atom before and after removing the linear regression term 
and  $L^2$-norm of forces, after removing data samples with force values unreasonably high are shown in Figure \ref{fig:hist_energy_forces}. 
The partition of each dataset into training, validation, and testing has been performed using the 80\%-10\%-10\% splitting. In Figure \ref{fig:hist_3set} we show the histograms of the distribution of energies and atomic forces for the training, validation, and testing subsets of each dataset. 

\subsection{Scalable data management}
\label{adios-io}

HydraGNN implements two optimization strategies that address scalability issues due to the large volume of data used for training.
These strategies aim for: 1) efficient storage and performant reading of large training data, and 2) fast reading of batch data during the training process. 
As atomistic materials modeling datasets are typically exported as collections of large numbers of files, storing datasets on a shared parallel file system (PFS) and then reading data from the large number of files during the training process causes a severe I/O bottleneck for GNN training. Multiple datasets cumulatively containing tens of thousands of small files put significant pressure on the PFS's metadata service, further slowing data access. Additionally, frequent data fetching by multiple GPUs from the PFS during training loops results in a substantial slowdown in the training process. We adopted a two-pronged approach to manage large data and reducing the I/O overhead for training the GNN model. 
First, we pre-process the various input datasets and store their graph representation using a scientific data management library. Secondly, we use a distributed in-memory data store to load data into memory for fast shuffling of data samples during the training process. 

\subsubsection{ADIOS for high performance I/O}

Several publicly available atomistic materials modeling datasets are stored using bespoke schemas and exported as large collections of files. For example, the \texttt{OC2020} dataset \cite{oc2020} consists of over 50,000 files. Storing multiple such datasets adds prohibitively high metadata overhead on the PFS and leads to slow data ingestion during the training process. 
For efficiently storing and reading large volumes of training data, we use the ADIOS ~\cite{godoy2020adios} scientific data management library, which provides a state-of-the-art solution for managing extreme-scale data.
ADIOS is designed to provide scalable I/O on the largest supercomputers in the world and has been successfully used in science applications that write and read several petabytes in a single simulation run.

An ADIOS file is stored in a hierarchical, self-documenting format that consists of a directory with sub-files and metadata files.
Data is stored in ADIOS variables and is automatically distributed across several files called ADIOS `sub-files.'
Users only focus on creating variables and issuing read/write calls, leaving the storage format and organization to ADIOS.
For example, we store graph node features in a large array which is automatically distributed amongst several sub-files when it is written to the ADIOS file. 
ADIOS internally maintains metadata to track the structure and organization of data.

The number of sub-files controls the concurrency level while reading data in parallel.
This $n:m$ pattern in which $n$ processes concurrently read data from $m$ sub-files is pivotal to obtaining high reading performance using ADIOS, which provides several options to tune I/O performance, including configuring the number of sub-files.
We create the graph structures from input data and store them in ADIOS as a separate pre-processing step.
We have developed a data writer and reader in HydraGNN for writing and reading graph data, respectively, from ADIOS files during the training process.
When an ADIOS file is created during the data pre-processing, we split the data samples into three groups - `trainset' representing training data, `valset' for data used for validation, and `testset' data for testing the model performance.
This logical grouping of data samples helps us read different groups of data samples for different tasks during the training process.

\subsubsection{DDStore}

\begin{figure*}
    \centering
    \includegraphics[width=0.99\textwidth]{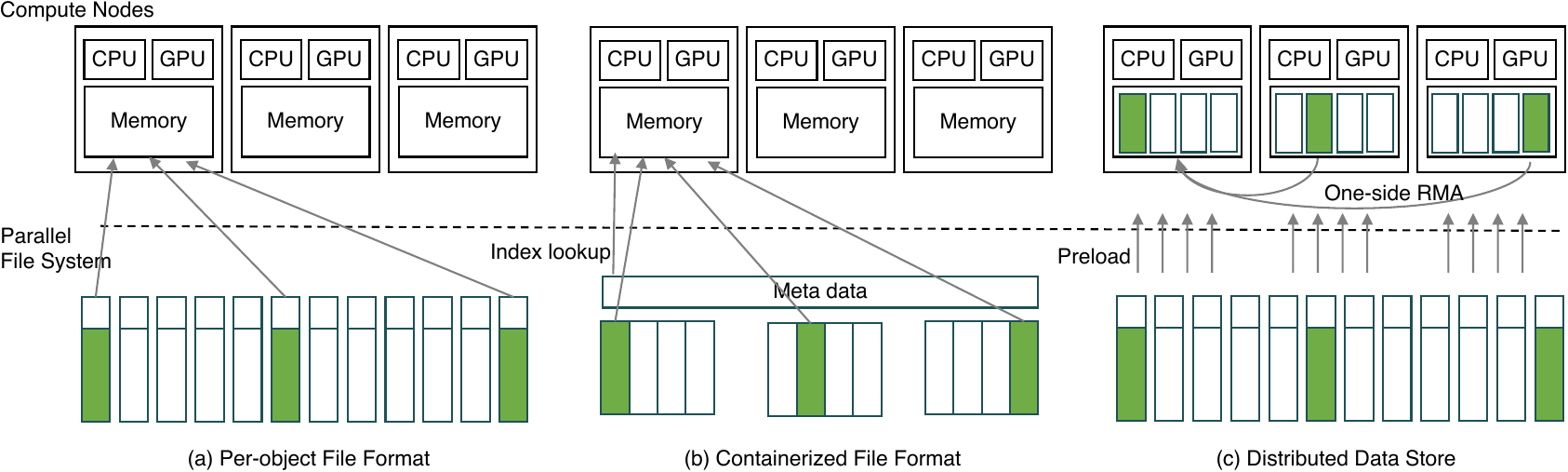}
    \caption{Different approaches for shuffling data during the training process. In a), data is read from the shared file system in which each graph object is stored in its own separate file (high file system metadata overhead). In b), data is read from an ADIOS file (low metadata overhead, high I/O bandwidth). In c), all data is read once into DDStore \cite{10.1145/3624062.3624171}, an in-memory data store which uses MPI one-sided RMA operations to obtain data from remote processes (best performance).}
\label{fig:ddstore}
\end{figure*}

DDP \cite{dean2008mapreduce, dean2012large, paszke2019pytorch, li2014scaling, goyal2017accurate} involves distributing training data amongst the available compute resources. 
Data is grouped into batches, and GPUs train on one batch at a time before fetching the next batch until all batches are processed in an epoch. Frequently reading data from the PFS, even via a high-performance library such as ADIOS, is an expensive operation because I/O over the shared PFS is one of the slowest operations in a computing system. 

To provide fast data retrieval during training, we used DDStore \cite{10.1145/3624062.3624171}, a distributed data store that provides in-memory data transfer between processes. 
When training begins, processes read data from ADIOS files and load into the node's memory, which maintains a global map of data samples on each process. When a GPU requests a new batch of data, DDStore fetches the data from remote processes using low latency, fast communication techniques rather than reading data from the PFS. 
By restricting access to the PFS to the initial bootup phase, DDStore ensures that obtaining a batch is a fast, in-memory operation. Experiments described in \cite{10.1145/3624062.3624171} show that it leads to a 6$\times$ speedup in overall training time.

DDStore provides options to tune the size of data chunks stored on each process (chunking), replicating a dataset on internal sub-groups of processes (replication), and the communication mechanism selected for fetching data. For our experiments, data is split evenly amongst all processes, and a single replica of the dataset is maintained across all processes. For efficient data retrieval, the low latency MPI one-sided remote memory access (RMA) operations were used. Fig. \ref{fig:ddstore} shows the data loading and caching approach used by DDStore compared to traditional approaches that read data directly from the file system.
Section \ref{performance_section} shows the time taken to obtain a batch of data samples for different model sizes and node counts.

\subsection{MTL for prediction of energy and atomic forces}
\begin{figure*}
    \centering
    \includegraphics[width=0.95\textwidth]{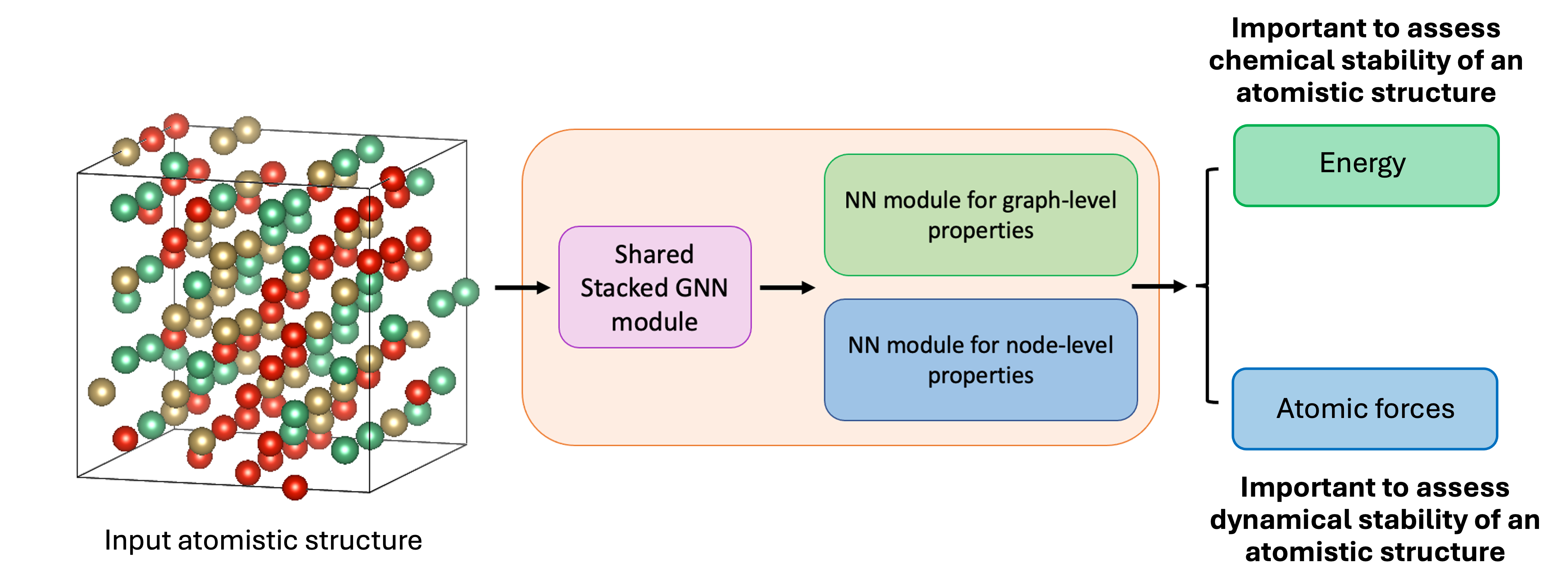}
    \caption{HydraGNN architecture for simultaneous prediction of energy and atomic forces.}
\label{fig:mtlgcnn}
\end{figure*}

MTL uses a single DL architecture to simultaneously predict multiple quantities \cite{caruana} and allows for a natural and automated incorporation of physics knowledge into the model by extracting correlations between the properties predicted, with manual intervention by a domain expert only needed in determining which quantities to use. The use of MTL is useful for developing GFMs, as it induces the model to use the physical correlations to develop physics-informed features that can be transferred on several downstream tasks. 
In this work, we choose energy and atomic forces as quantities to predict simultaneously. Denoting the number of atoms in an atomistic structure with $n$ and the Cartesian coordinates of the position of the nuclei of atom $i$ (where $i$ can be any integer between 1 and $n$) with $\mathbf{x}_i\in \mathbb{R}^3$, the relation between the energy $e \in \mathbb{R}$ and forces $\mathbf{f}_i \in \mathbb{R}^3$ acting on atom $i$ is
\begin{equation}
\mathbf{f}_i = - \nabla_{\mathbf{x}_i} e.
\end{equation}
In HydraGNN, each predicted quantity is associated with a separate loss function and the global objective function minimized during the training is a linear combination of the individual loss functions. We denote by $\mathbf{F}\in \mathbb{R}^{n\times 3}$ the atomic force tensor that contains the atomic forces acting on all the $N$ atoms
\begin{equation}
\mathbf{F} = 
\begin{bmatrix}
\bold{f}^T_1 \\ \vdots \\ \bold{f}^T_N \end{bmatrix} ,
\label{hat_f}
\end{equation}
where $T$ represents the transposition of a vector. 
Denoting by $P$ the total number of parameters to train in the GFM architecture, the MTL loss function $\ell_\mathrm{MTL}:\mathbb{R}^{P}\rightarrow \mathbb{R}^+$ is:
\begin{align}
 \ell_\mathrm{MTL}(\mathbf{W}) =  \alpha_\mathrm{energy} \lVert e_{\textrm{predict}}(\mathbf{W}) 
    - e \rVert_1 + \alpha_\mathrm{forces} \lVert \mathbf{F}_{\textrm{predict}} (\mathbf{W})
    - \mathbf{F} \rVert_1,
\end{align}
where $\lVert \cdot \rVert_1$ represents the $L_1$ norm of a vector or the induced tensor norm, $e$ and $\mathbf{F}$ represent the true values of energies and forces, $\mathbf{e}_{\textrm{predict}}$ and $\mathbf{F}_{\textrm{predict}}$ are the corresponding predictions given by HydraGNN, and the scalar positive weights $\alpha_\mathrm{energy}$ and $\alpha_\mathrm{forces}$ are used to calibrate the individual terms of the loss function for each individual properties. Similarly to \cite{batzner2024}, we calibrate the scalar positive weights $\alpha_\mathrm{energy}$ and $\alpha_\mathrm{forces}$ to account for the fact that the energy is a global quantity, while the atomic forces are local quantities. More specifically, we set the values of these scalar weights to $\alpha_\mathrm{energy}=1$ and $\alpha_\mathrm{forces}=100$. An illustration of the main components of the HydraGNN architecture used for MTL is described in Figure~\ref{fig:mtlgcnn}.

\subsection{Scalable HPO} 
GNNs are known for their exceptional performance in learning from graph-structured atomistic materials modeling datasets. However, their development and broader application are hindered by the need for meticulous tuning of the network architecture. To achieve high predictive accuracy across chemically diverse datasets, it is essential to fine-tune the hyperparameters of HydraGNN. 
The task of identifying optimal hyperparameter settings is daunting and has been extensively documented in existing literature \cite{LUPOPASINI2021102788, Yu2020HyperParameterOA, Muyskens2021MuyGPsSG, Kadra2023ScalingLF, fetterman2023tune, NIPS2011_86e8f7ab}. Manual tuning requires extensive experimentation and often results in suboptimal performance. 

To perform HPO at large scale, we used DeepHyper \cite{deephyper}, an open-source Python package designed for optimizing hyperparameters, searching for optimal neural architectures. 
Specifically, we used asynchronous Bayesian optimization that continuously refines a surrogate model by sampling hyperparameter configurations. 
The efficacy of DeepHyper's asynchronous Bayesian optimization has been demonstrated across various DL benchmarks, outperforming methods such as random search, genetic algorithms, and Hyperband in environments equipped with CPUs and GPUs.
In the DeepHyper setup, a manager node refines the surrogate model and suggests promising configurations while worker nodes perform the evaluations. Our approach uses a centralized architecture with process-based parallelism, optimizing the allocation of tasks across computing nodes to avoid bottlenecks. 

Message passing is the core methodology of GNN models since it prescribes how features of nodes and edges are updated using information contained in neighboring nodes and edges. Various MPNNs have been developed and tailored for different atomistic systems, such as SchNet \cite{10.5555/3294771.3294866} for organic molecules and CGCNN \cite{PhysRevLett.120.145301} for solid state crystals. However, when considering GFMs applicable to a broad range of systems in atomistic materials, it is not practical to confine ourselves to a specific MPNN method. In HydraGNN, the choice of MPNN is configurable through a hyperparameter, allowing the users to select the optimal model that best suits their applications. 
We include MPNN as a categorical hyperparameter in the HPO runs to allow for the identification of the best performing MPNN layers for the assigned training data. 

In HPO, early termination strategies are vital for improving the utilization of computational resources by discarding unpromising candidates based on their performance trends.
This decision has proven effective early in the training process \cite{egele2024}. DeepHyper provides three early discarding techniques suited for asynchronous and parallel environments: (1) asynchronous successive halving, which progressively eliminates candidates based on their interim performance; (2) learning curve extrapolation, which predicts future performance from early data and facilitates early termination; and (3) constant fidelity, which sets a fixed resource allocation for each candidate before deciding whether to continue. 
For our tests, we used constant fidelity as it enables efficient reallocation of resources towards more promising configurations and significantly enhances operational efficiency in large-scale, distributed computing environments. 
We used 10 epochs as a stopping criterion for each model training in the HPO phase.
While HPO has been previously explored for GNNs, our approach uses HPO on a scale previously unattempted. 



\subsection{Scalable UQ with GNN Ensembles}
Ensemble methods are widely utilized in UQ to compile predictions from various models, termed ensemble members, into a unified forecast. The goal of these methods is to enhance model generalization by drawing on the diverse capabilities of each individual model \cite{krogh1994}. To promote a varied set of predictions, practices such as different model initializations, techniques like Bagging and Boosting, and the integration of diverse network architectures are used. Research conducted by Egele et al. \cite{9956231} showed that expanding the variety of network architectures within an ensemble can improve the diversity, thereby increasing the precision of uncertainty assessments. They also developed a technique for concurrently training multiple candidate models, which optimizes the use of computational resources. Ensemble methods are acknowledged for their ability to deliver reliable uncertainty estimates and their ease of implementation and scalability, making them practical for various UQ applications.

To account for model (epistemic) uncertainty, we employ ensembles consisting of multiple neural networks (NNs). Our approach involves considering a collection of GNN models generated by DeepHyper, denoted by $\mathcal{C} = \{\theta_i, i=1,2,\cdots,c\}$. We then select $K$ models from this collection to form the ensemble, where $\mathcal{E} = \{\theta_i, i=1,2,\cdots,K\}$ and $K$ denotes the ensemble size. For an input graph $G$, the ensemble's prediction is the average of prediction from all model members $f_{\theta_i}$, 
\begin{equation}
    \Tilde{y}=\frac{1}{K}\sum_{i=1}^K f_{\theta_i}(G),
\end{equation}
and the uncertainty is measured as the standard deviation (STD),
\begin{equation}\label{eq-enstd}
    \sigma_{\Tilde{y}}=\sqrt{\frac{1}{K}\sum_{i=1}^K \left(f_{\theta_i}(G)-\Tilde{y}\right)^2}.
\end{equation}

Our method offers notable advantages in terms of generality and scalability. Central to our approach is the construction of model ensembles, which relies on scalable HPO. This methodology can be applied to any type of NN model. The process begins with using a standard NN architecture, conducting HPO, selecting the most suitable models, and subsequently producing uncertainty estimates. The scalability of our method is anchored in both the scalable nature of the hyperparameter search and the ability to train ensembles efficiently.  Working with an ensemble of models enables many options for building consensus models, uncertainty estimation, and active learning~\cite{krogh1994}.




\section{Performance measurements}
\label{performance_section}



\subsection{Setup}
\label{performance_setup}

For strong and weak scaling studies we utilize three different sizes of GFM architectures, denoted as SMALL, MEDIUM, and LARGE. They differ in the total number of parameters, ranging from approximately 60,000 to 163 million. Table~\ref{tab:modelsize} provides details about the three model sizes.

Experiments were conducted on two DOE supercomputers: Frontier at ORNL and Perlmutter at NERSC. Both systems provide state-of-the-art GPU-based heterogeneous architectures.
Frontier, located at the Oak Ridge Leadership Computing Facility at ORNL, is currently the world's fastest supercomputer~\cite{top500}. It comprises a total of 9,408 compute nodes, each featuring a single 64-core AMD EPYC 7763 (Milan) CPU and four AMD Instinct MI250X GPU accelerators, effectively providing eight GPU units per node. Running with one rank per GPU unit, each rank has 64 GB of DDR4 (CPU) and 64 GB of HBM2e (GPU) memory.

Perlmutter, a supercomputer at National Energy Research Scientific Computing Center (NERSC), features approximately 3000 CPU-only nodes and 1800 GPU-accelerated nodes. Our work utilizes only the GPU-accelerated nodes. Each node is equipped with an AMD EPYC 7763 CPU and four NVIDIA Ampere A100 GPUs interconnected via NVLink-3. Running with one rank per GPU unit, each rank has 64 GB of DDR4 (CPU) and 40 GB of HBM2 (GPU) memory. Both Frontier and Perlmutter use HPE Cray Slingshot$^\mathrm{(TM)}$ interconnects.



To aid in monitoring HydraGNN execution in real-time for a subset of the analysis carried out on Frontier, an AMD Research utility, {\em Omnistat}~\cite{omnistat}, was used to sample a variety of GPU telemetry metrics including occupancy, high-bandwidth memory (HBM) usage, power, temperature, energy consumption, and clock/memory frequencies on a per GCD basis across all nodes assigned to an individual run. This Python-based utility was executed entirely in user-space implemented as a Prometheus client on each assigned compute node and combines low-overhead sampling via AMD's system management interface (SMI) at fixed intervals with a temporary Prometheus server~\cite{prometheus_2015} instantiated on one CPU core of the master compute host per batch job.  Minimal job overhead (less than 0.5\%) was observed when running HydraGNN training with this approach for sampling intervals down to one second.


\begin{table*}[tb]
\normalsize
  \centering
  \begin{tabular}{|l|r|r|r|}
  \hline
\textbf{Model size} & SMALL & MEDIUM & LARGE \\
\hline
Type of MPNN layer & EGNN & EGNN & EGNN \\
\hline
\# MPNN layers & 3 & 6 & 6\\
\hline
\# neurons in MPNN layers & 50 & 500 & 2,000\\
\hline
\# FC layers & 2 & 2 & 3\\
\hline
\# neurons in FC layers & 50 & 1,000 & 1,000 \\
\hline
\textbf{Number of parameters} & \textbf{58,404} & \textbf{14,539,004}  & \textbf{163,129,004} \\
  \hline
  \end{tabular}
  \vspace*{0.1in}
  \caption{GNN model sizes used for strong and weak scaling tests on NERSC-Perlmutter and OLCF-Frontier.}
  \label{tab:modelsize}
\end{table*}

\subsection{I/O performance for reading large data}
In Section \ref{adios-io}, we described using the ADIOS scientific data management library for fast storage and retrieval of large training data. In this section, we show the performance of reading large data in HydraGNN for training models.

Of all the datasets used in this study, the Open Catalyst 2020 dataset is the largest in terms of number of atomistic structures, storage size of the dataset, and number of files across which data is stored.
The original dataset consists of over 50,000 files.
The dataset was pre-processed into ADIOS and was configured to use just over 50 ADIOS sub-files, which led to a 1000$\times$ reduction in the metadata footprint.

When training begins, HydraGNN reads ADIOS data in parallel on all processes.
This read operation is a two-step process in which first the root process obtains the number of graphs (atomistic structures) followed by the size (number of atoms) and the feature metadata for each graph.
This information is broadcast to all other processes that implicitly distribute the graphs evenly amongst themselves and concurrently read their assigned graphs from the ADIOS file.
This set of operations is performed for all atomistic structures groups - training set, validation set, and testing set.

\begin{figure}[ht!]
    \centering
    \includegraphics[width=0.6\textwidth]{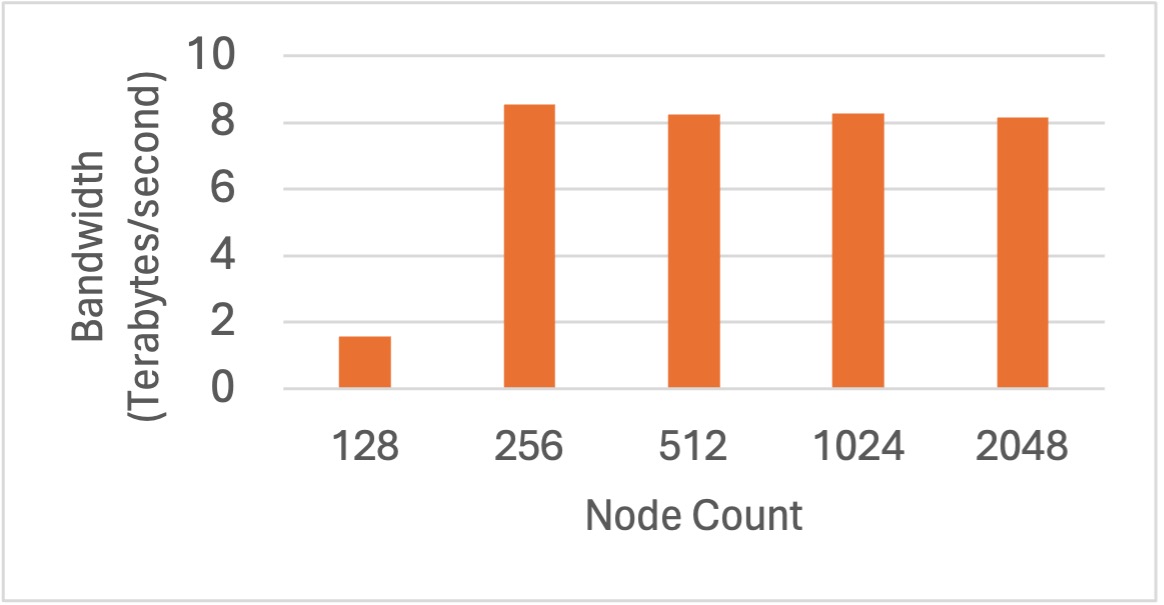}
    \caption{Read performance for the ADIOS `trainset' data of the \texttt{OC2020} dataset on Frontier. We obtain over 8 Terabytes/second for almost all node counts for reading 3.8 Terabytes of trainset data.}
\label{fig:adios-trainset}
\end{figure}

Fig. \ref{fig:adios-trainset} shows the reading performance of the training data on Frontier when all processes read their assigned graphs in parallel. We obtain over 8 Terabytes/second for higher node counts and almost 2 Terabytes/second on 128 nodes. The high I/O bandwidth is a characteristic feature of the ADIOS library as it permits multiple processes to read data spread over multiple ADIOS sub-files efficiently. A similar run on Perlmutter was not possible due to PFS issues encountered on the system during our study.

\begin{figure}[ht!]
    \centering
    \includegraphics[width=0.6\textwidth]{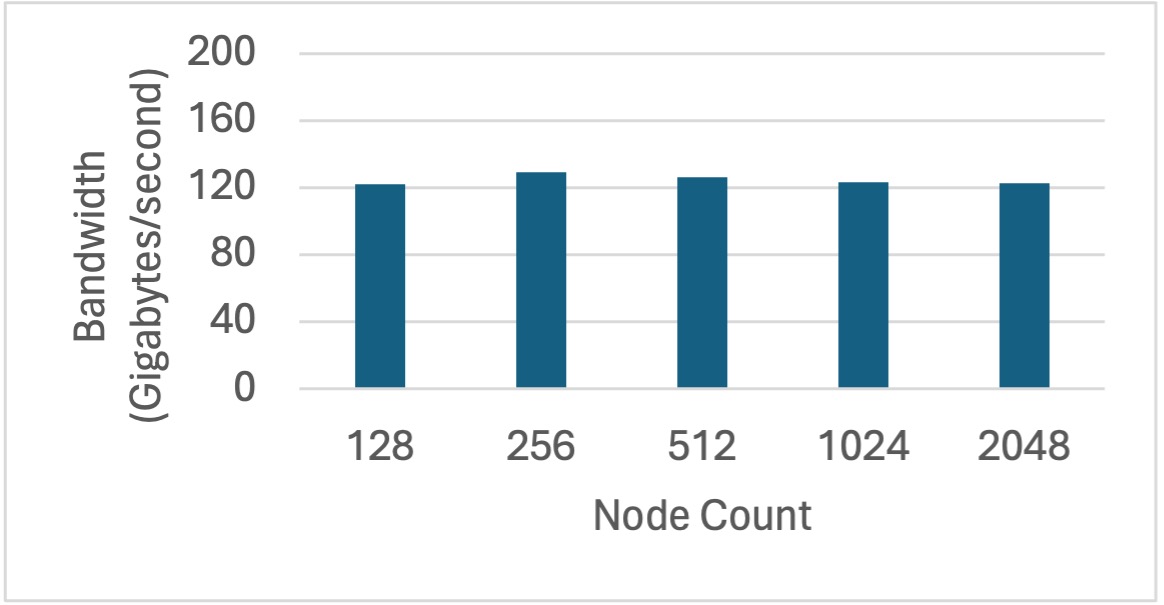}
    \caption{Read performance for the entire \texttt{OC2020} dataset on Frontier that includes training, validation, and testing data. We obtain over 120 Gigabytes/second (approximately 35 seconds) for reading 4.3 Terabytes of data. We will apply techniques to read user metadata in parallel to improve the overall read performance of the complete dataset.}
\label{fig:adios-full}
\end{figure}

The initial step in which the root process reads several small portions of the dataset and broadcasts them is an inherently sequential set of operations. As this slows the overall I/O, we obtain lower I/O bandwidth as the root process performs these tasks for the trainset, valset, and testset data groups to read a total of approximately 500 Gigabytes of initial data.
Fig. \ref{fig:adios-full} shows the sustained I/O bandwidth achieved when HydraGNN reads the entire \texttt{OC2020} dataset, which is 4.3 Terabytes in size. We obtain a net bandwidth of over 120 Gigabytes/second on Frontier, which allows HydraGNN to ingest the full collection of 120 million graphs in just over 30 seconds.

\subsection{HydraGNN Training Scaling Results}
We now analyze the scaling performance of HydraGNN on Frontier and Perlmutter.
We present weak and strong scaling trends, along with a breakdown of component operations in HydraGNN as we scale it up.
Experiments were performed with up to 2,048 nodes on Frontier and 256 nodes on Perlmutter
using the three model sizes discussed in Table \ref{tab:modelsize}.

\subsubsection{Weak scaling}

\begin{figure}[ht!]
    \centering
    \includegraphics[width=0.8\textwidth]{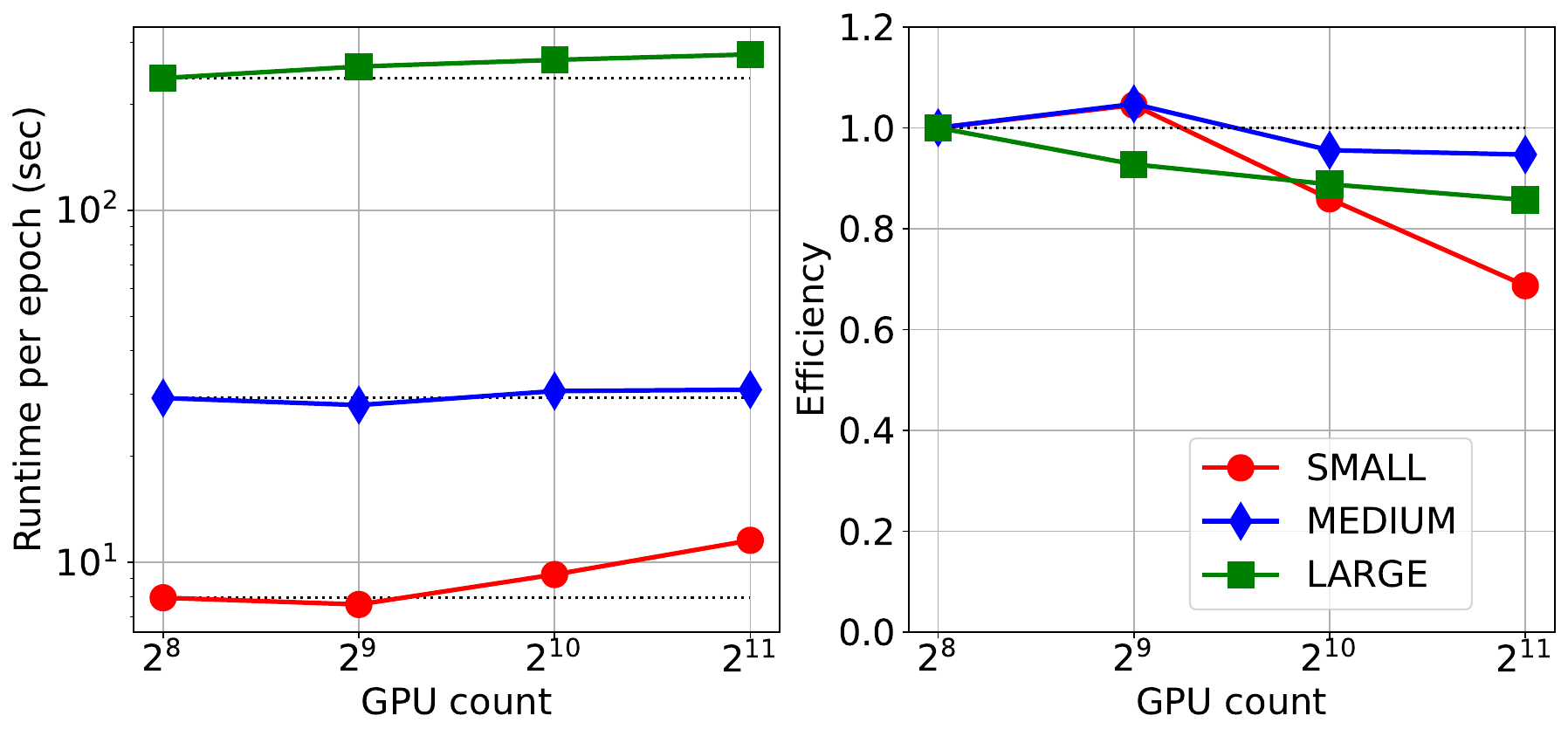}
    \includegraphics[width=0.8\textwidth]{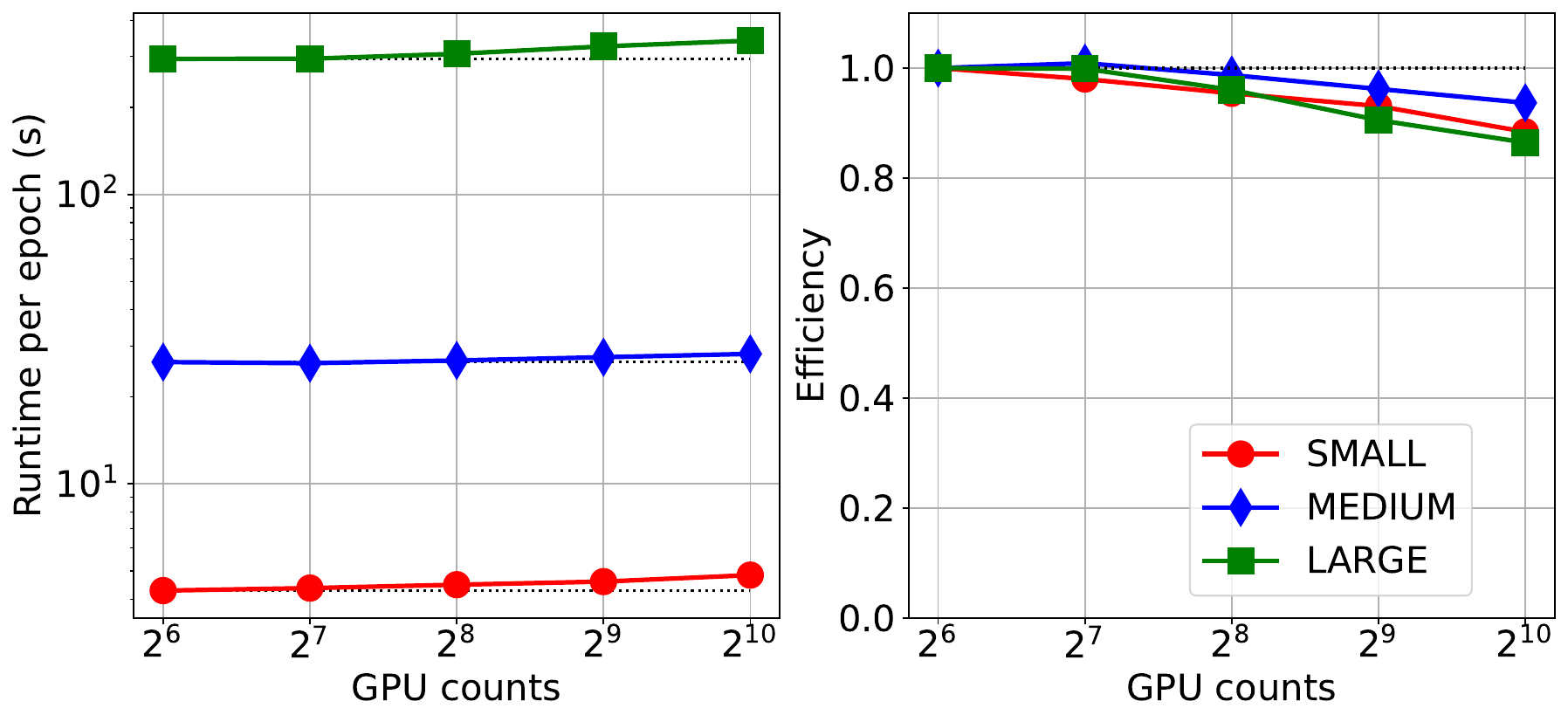}
    \caption{Weak scaling of HydraGNN multitasking pre-training on a problem of size 3,500 atomistic structures per GPU for (top) Frontier and (bottom) Perlmutter.}
\label{fig:weakscaling}
\end{figure}

For the weak scaling runs, we configured each GPU to process 3,500 atomistic structures equally. Fig. \ref{fig:weakscaling} shows the weak scaling performance on Perlmutter and Frontier as we vary the number of GPUs used for the training.
The reported time represents the average training time per epoch.
We conducted experiments with up to 2,048 GPUs on Frontier and 1,024 GPUs on Perlmutter. The number of GPUs on Perlmutter was dictated by constraints on available node hours.
We observe that the parallel efficiency of weak scaling experiments drops as we increase the number of GPUs beyond 256 for both Perlmutter and Frontier. This is attributed to increased communication costs as we scale the number of GPUs and the overhead associated with using varying graph sizes.


\begin{figure}[ht!]
    \centering
    \includegraphics[width=0.8\textwidth,keepaspectratio]{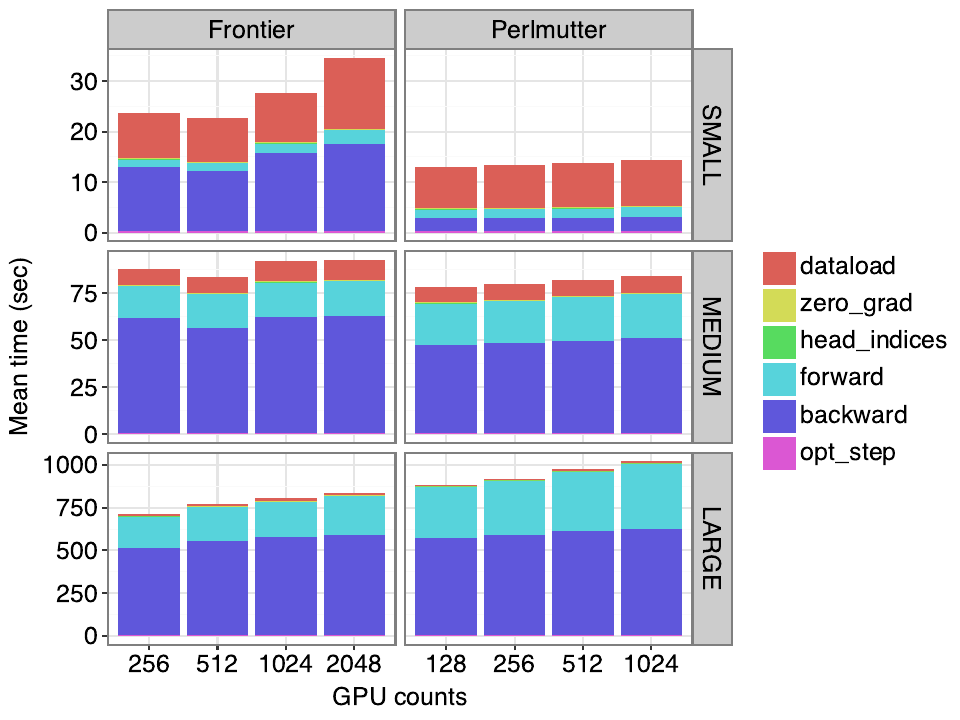}
    \caption{Weak scaling. AMD is better for large models.}
\label{fig:weakdetail}
\end{figure}

Fig. ~\ref{fig:weakdetail} provides a breakdown of the overhead of different components of HydraGNN used in the weak scaling experiments.
The terms `forward' and `backward' represent the forward and backward phases of the DL model training, respectively, and `dataload' denotes the cost of obtaining the next batch of data samples from DDStore after a GPU finishes processing its current batch.
We notice that `dataload' has a fixed cost, which expectedly becomes more prominent for the small model size and is only a fraction of the runtime as the model size increases. The forward and backward phases show an increase in runtime as we scale up the workflow as synchronization and communication operations become more expensive with increasing GPU counts.

\subsubsection{Strong scaling}
\begin{figure}[ht!]
    \centering
    \includegraphics[width=0.48\textwidth]{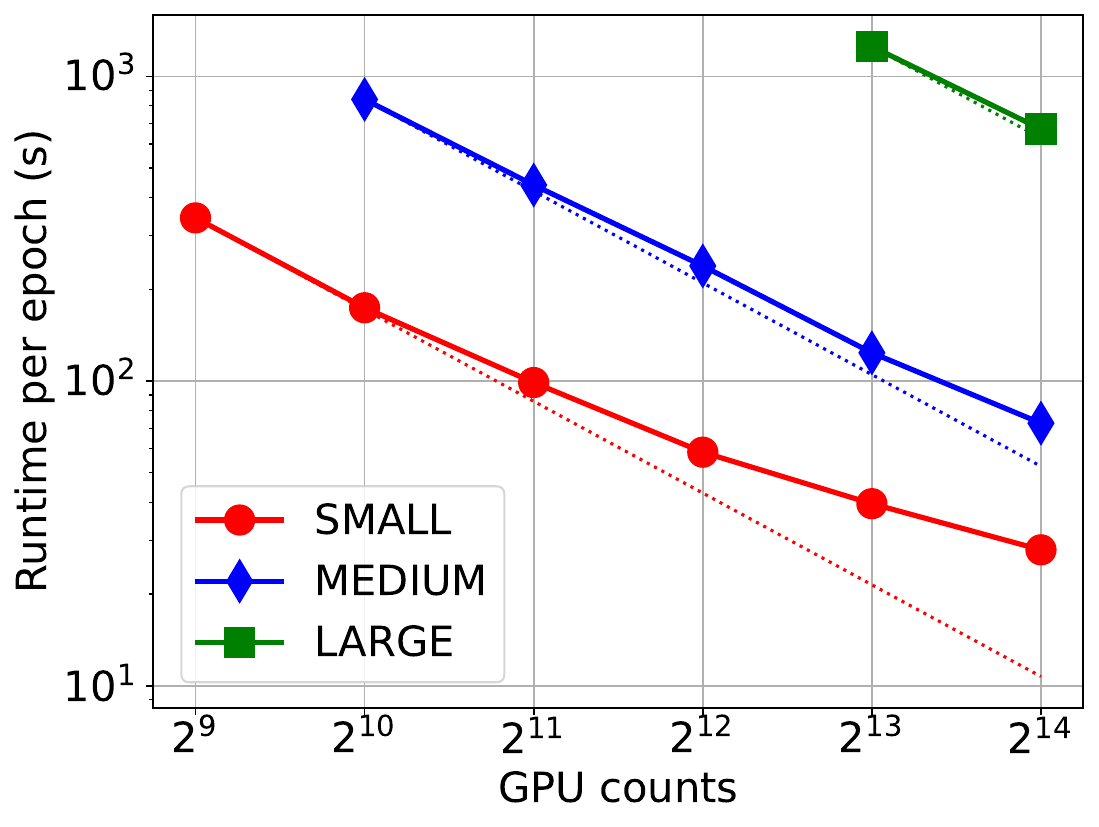}
    \includegraphics[width=0.48\textwidth]{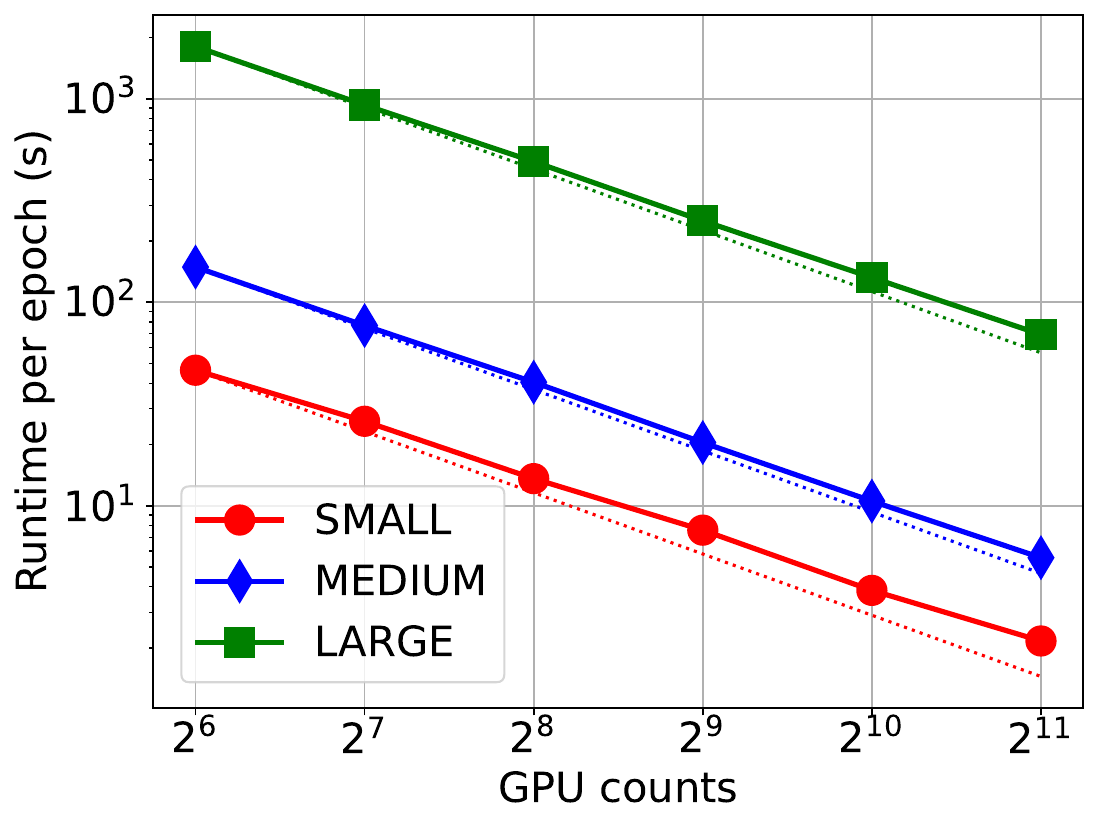}
    \caption{Strong scaling of HydraGNN multitasking pre-training on a problem of 120 million graphs on Frontier and 2 million graphs on Perlmutter with three GNN model sizes.}
\label{fig:strongscaling}
\end{figure}

For strong scaling runs, we trained HydraGNN on a subset of the entire data available, comprising 120 million atomistic graphs (approximately 4TB in size), on Frontier and 2 million atomistic graphs on Perlmutter for the three model sizes.
Fig. \ref{fig:strongscaling} shows the scaling results for 512 to 16,384 GPUs on Frontier and from 64 up to 2,048 GPUs on Perlmutter.
The reported time is the average training time per epoch, similar to the weak scaling measurements.
While the SMALL model's performance deviates from the optimal linear dotted line after 2,048 GPUs on Frontier, the MEDIUM and LARGE models maintain close to linear scaling up to 16,384 GPUs on Frontier.
We notice a similar trend on Perlmutter where we observe near-linear scaling up to 2,048 GPUs for all model sizes.

The drop in scaling performance is attributed to load imbalance - an artifact of small graph sizes.
As shown in Fig. \ref{fig:hist_atoms_edges}, we use a diverse dataset where graph sizes vary by up to 400 nodes, and the number of edges in the larger graphs exceeds 12,500. This results in an imbalanced workload among GPUs in each batch, causing some GPUs to finish training before others. As GPUs must synchronize for exchanging model weights, the runtime is dominated by the GPUs that must train on larger graphs. Effectively, this leads to sub-standard utilization of compute resources and poses a challenge towards achieving high-performant, scalable training.
Thereofre, while training on large volumes of data can help develop robust models because of the diverse nature of data, the computational performance may suffer as the workload can vary greatly.

\begin{figure}[ht!]
    \centering
    \includegraphics[width=0.48\textwidth]{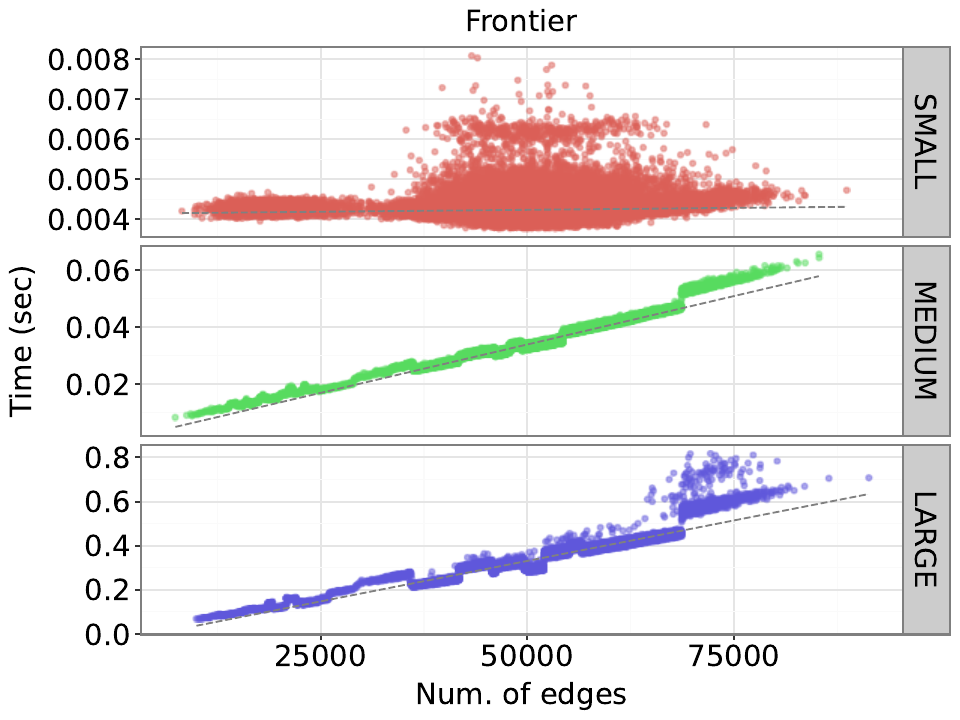}
    \includegraphics[width=0.48\textwidth]{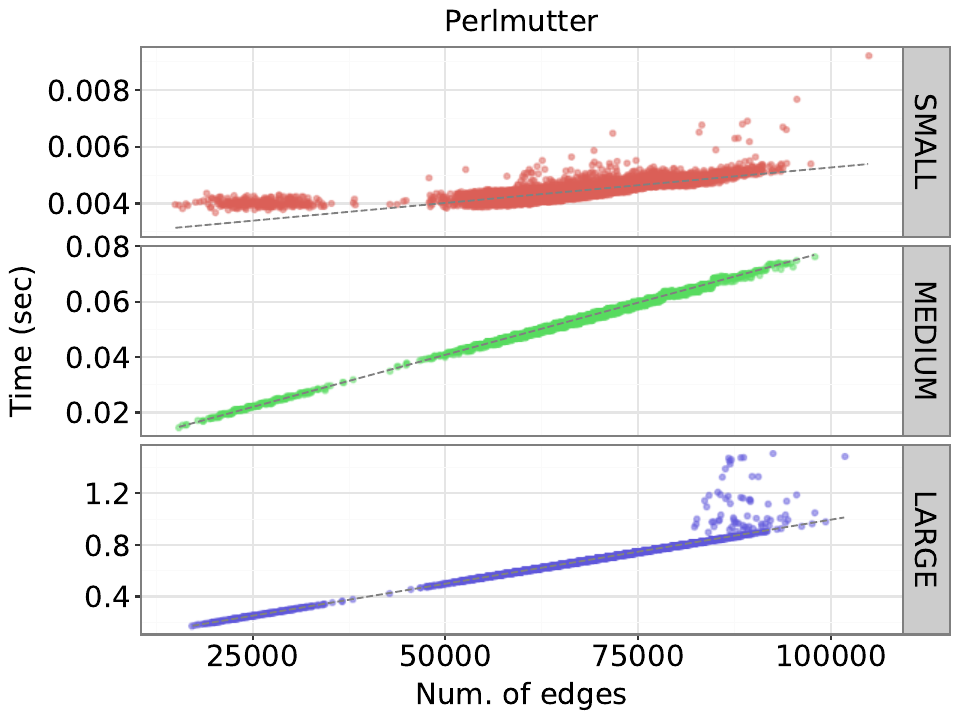}
    \caption{The distribution of forward time during the training of three models with respect to the graph size, measured in the number of edges. It illustrate a linear relationship between forward time and graph size.}
\label{fig:scatter}
\end{figure}

\begin{figure}[ht!]
    \centering
    \includegraphics[width=0.6\textwidth]{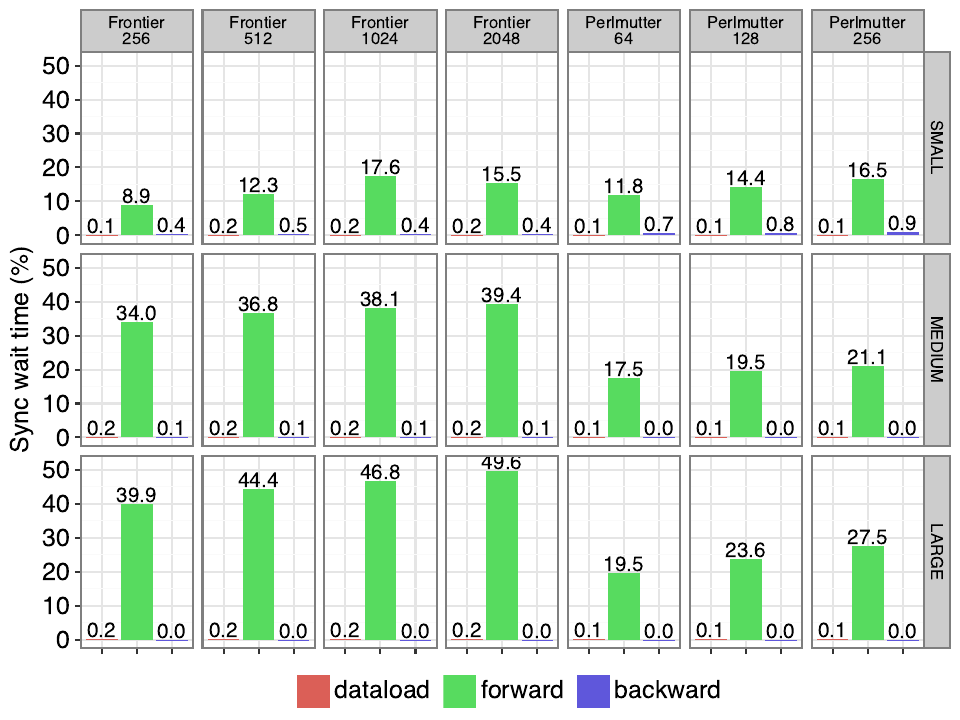}
    \caption{Average percentage of waiting time during three parallel tasks -- data loading, forward pass, and backward pass.}
\label{fig:syncwait}
\end{figure}

The EGNN model we used is particularly vulnerable to this problem. The time required for forward calculations in EGNN is directly proportional to the number of edges in the graph sample. For datasets with highly variable edge counts between graph samples, the likelihood of load imbalance between GPUs increases. 
Fig.~\ref{fig:scatter} illustrates the time spent on the forward task in the EGNN model with different model sizes. We observe an almost linear relationship between forward time and graph size at each batch (measured by the number of edges). The SMALL models show large variances on both machines, which is expected due to system noise being more consequential for smaller model sizes. Significant performance differences (e.g., the difference between minimum and maximum time) are observed due to the varying graph sizes in our datasets. However, for other tasks (data loading and backward), we do not observe a similar correlation, as they are agnostic of the graph size.
Fig.~\ref{fig:syncwait} illustrates the average percentage of time spent waiting during three tasks: data loading, forward pass, and backward pass. It highlights a significant waiting period during the forward pass, primarily due to varying graph sizes. This waiting time increases as the disparity in graph sizes among GPUs grows. Other tasks, such as data loading and backward pass, also involve waiting time, but to a much lesser extent.

\begin{figure}[ht!]
    \centering
    \includegraphics[width=0.6\textwidth]{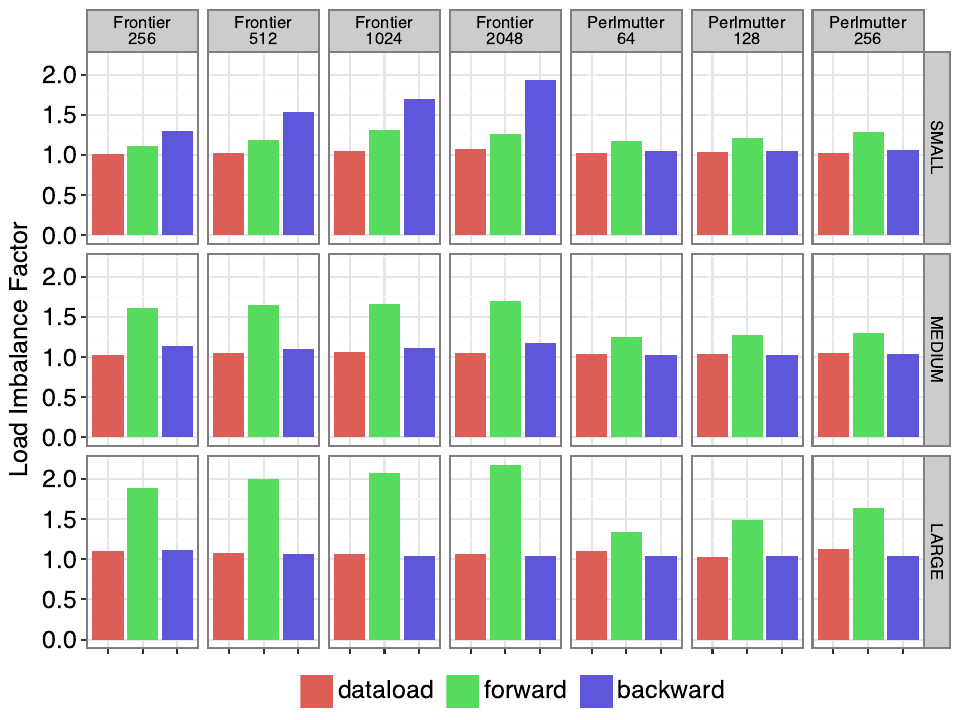}
    \caption{Load imbalance factor.}
\label{fig:imbalancefactor}
\end{figure}

To quantify the degree of load imbalance between GPUs, we compute the load imbalance factor (LIF) defined by the ratio
\begin{equation}
    LIF = T_{max} / T_{avg}
\end{equation}
where $T_{max}$ and $T_{avg}$ represent the maximum runtime and the average runtime for training an epoch, respectively, among all computing resources (GPUs in our case).
These times represent the time to perform training (forward and backward calculations) and do not include wait times during synchronization.
For a well-balanced workload, LIF approaches 1.0 from above, whereas it increases as the workload imbalance increases.
Fig.~\ref{fig:imbalancefactor} presents the LIF scores that show the imbalance among processes. The trend remains consistent: while data loading and backward pass exhibit nearly balanced workloads (with scores close to 1.0), the forward pass shows imbalanced workload characteristics as it deviates from 1.0.

To address the performance penalties caused by workload imbalance, one potential solution is to implement binning or sharing approaches based on graph sizes. This would help ensure balanced workloads across multiple GPUs during each batch processing. However, there is a concern that this method might negatively impact the quality of training or the training losses during the optimization phase by reducing the stochastic effect, which is crucial for effective training. 
Given this potential trade-off, it is crucial to explore and develop more sophisticated strategies to mitigate load imbalance while maintaining training quality. This will be a key focus for future research and development efforts.

\subsection{Scalable HPO}

The HPO process is performed using DeepHyper \cite{deephyper}. DeepHyper has been specifically designed to perform efficient and scalable HPO on integrated extreme scale HPC and leadership class supercomputing facilities, and it thus suits well our purpose. Among the various hyperparameter search algorithms implemented in DeepHyper, we used the Centralized Bayesian Optimisation Search \cite{Jones1998, Swersky2013, Snoek2012, Snoek2014, Gonzalez2016, Frazier2018}, previously named as “Asynchronous Model-Based Search” (AMBS) \cite{Hutter2012}. It follows a manager-workers architecture where the manager runs the Bayesian optimization loop and workers execute parallel evaluations of the black-box function.

The hyperparameter tuning has spanned important architectural hyperparameters described in Table \ref{tab:hyperparameter}. The range of architectural hyperparameters covers regions of the hyperparameter space that allow to construct HydraGNN models of extremely diverse size, which include the SMALL model and the LARGE models described in Table \ref{tab:modelsize} as extremes.

\begin{table*}[ht!]
\normalsize
\centering
\begin{tabular}{|l|r|r|}
\hline
\textbf{Hyperparameter} & \textbf{Type} & \textbf{Admissible values} \\
\hline
 Type of MPNN  layer &  Categorical &  \{PNA, EGNN, SchNet\} \\
 \hline
\# MPNN layers & Integer & \{1,\ldots,6\} \\
\hline
\# neurons in MPNN layers & Integer & \{100, \ldots, 2,000\} \\
\hline
\# FC layers & Integer & \{2,3\}\\
\hline
\# neurons in FC layers & Integer & \{300, \ldots, 1,000\}\\
\hline
\# batch size & Integer & \{16, \ldots, 128\}\\
\hline
\end{tabular}
\vspace{0.1in}
\caption{Set of architectural HydraGNN hyperparameters tuned by scalable HPO. }
\label{tab:hyperparameter}
\end{table*}

We ran four consecutive HPO runs of progressively increasing scale, each restarting from the output of the previous one. During each one of the four HPO runs, each HPO trial is associated with an independent `srun' execution of the SLURM scheduler and occupies 128 Frontier nodes (i.e., 1,024 AMD Instinct MI250x GCDs) for distributed training using DDP. Concurrent HPO trials are executed asynchronously, and the termination of an HPO trial is immediately followed by the start of a new one on the same set of compute nodes.  Throughout the entire execution of each HPO run, we used the open-source Omnistat data-collection infrastructure highlighted in Section~\ref{performance_setup} to simultaneously collect GPU telemetry with measurements saved on a local Lustre PFS. Since each HPO trial is submitted as a separate job step within the global job submission, Omnistat post-processing allows us to isolate power and energy measurements at the job-step level thus providing an aggregate GPU energy consumed estimate for each trial independently.  In order to ensure that the HPO process is performed in an energy-efficient way on OLCF-Frontier, we early stop the training of HydraGNN models for each HPO trial after 10 epochs. This number of epochs allows to early stop the HPO trials that are clearly underperforming in a timely manner, without wasteful energy consumption caused by further training epochs that would not likely improve their accuracy, while still ensuring that promising HPO trials are distinguishable and selected for the next computational tasks. This approach results in impactful energy savings.
Our use of DeepHyper for asynchronous Bayesian optimization, combined with in-band telemetry monitoring and a strategic deployment of early termination strategies, showcases a significant advancement in the field, optimizing GNN training in ways that have not been documented prior to this work. 

The first HPO run used 2,048 Frontier nodes in parallel, thereby allowing 16 distinct HydraGNN architectures to be concurrently trained. This first HPO run allowed us to perform a first uninformed exploration of the hyperparameter space, and construct some guidance for the consecutive HPO runs. The results of the first HPO run have been used as inputs to inform the second HPO run, which used 3,072 Frontier nodes in parallel, thereby allowing 28 distinct HydraGNN architectures to be concurrently trained. 
The results of the second HPO run have been used as input to inform the third HPO run, which use 4,096 Frontier nodes in parallel, thereby allowing 32 distinct HydraGNN architectures to be concurrently trained. 
Similarly, the output of the third HPO run has been used to successfully guide the fourth and last HPO run, which used 8,560 Frontier nodes in parallel, thereby allowing 67 distinct HydraGNN architectures to be concurrently trained. 
This extensive scale of the last HPO run not only tests the limits of scalability and efficiency in computational resources, but also addresses the challenges associated with the high dimensionality of the hyperparameter space that needs to be explored, ensuring that a judicious balance between exploitation and exploration is maintained.

\begin{figure}[ht!]
    \centering
        \includegraphics[width=0.6\textwidth]{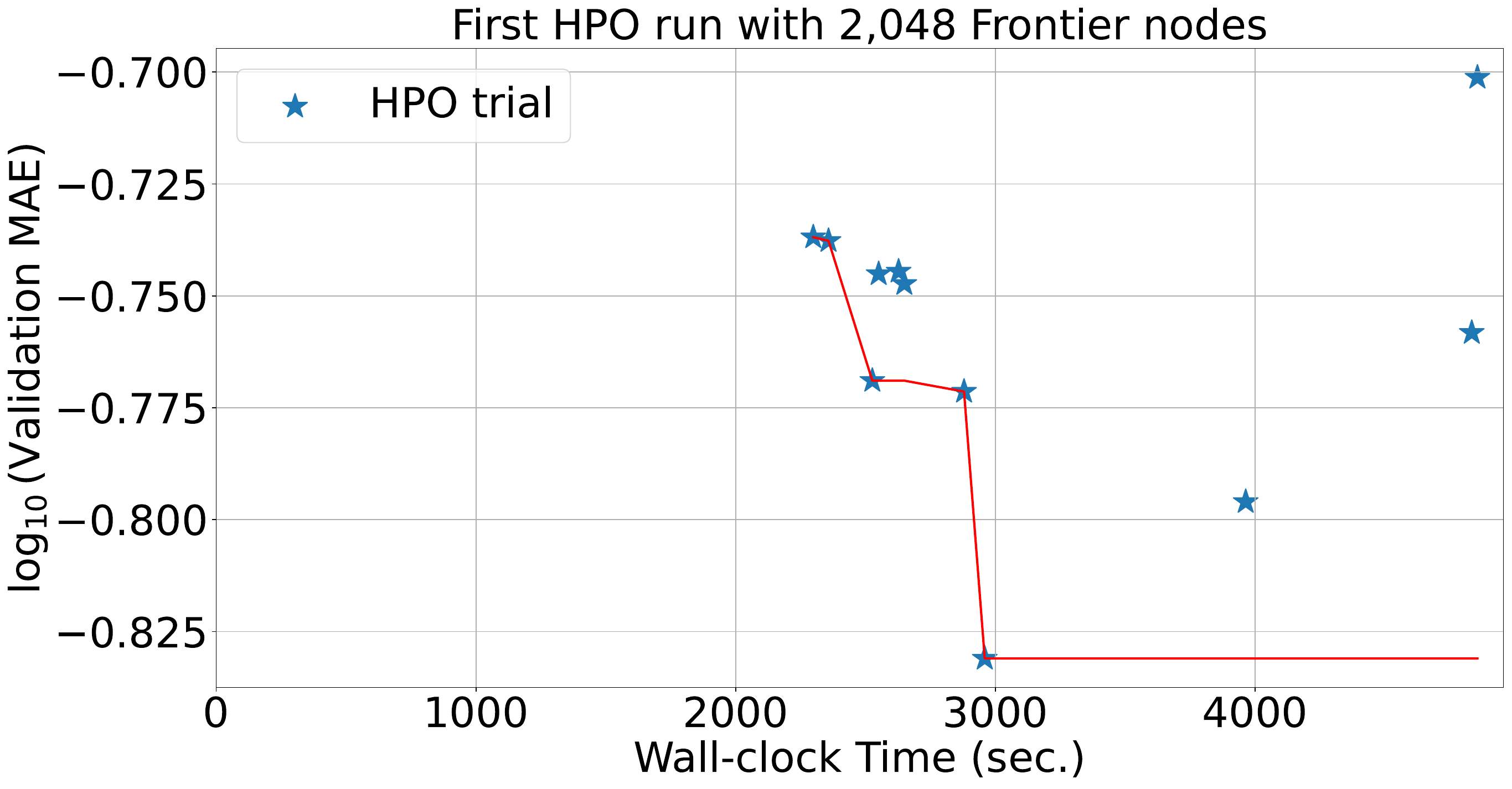}
        \caption{Logarithm in base 10 of validation MAE for HPO trials of the first HPO run on 2,048 Frontier nodes as a function of wall-clock time expressed in seconds. The red line shows the cumulative minimum across wall-clock time. }
        \label{fig:hpo_objective_1}
\end{figure}

\begin{figure}[ht!]
        \centering
        \includegraphics[width=0.6\textwidth]{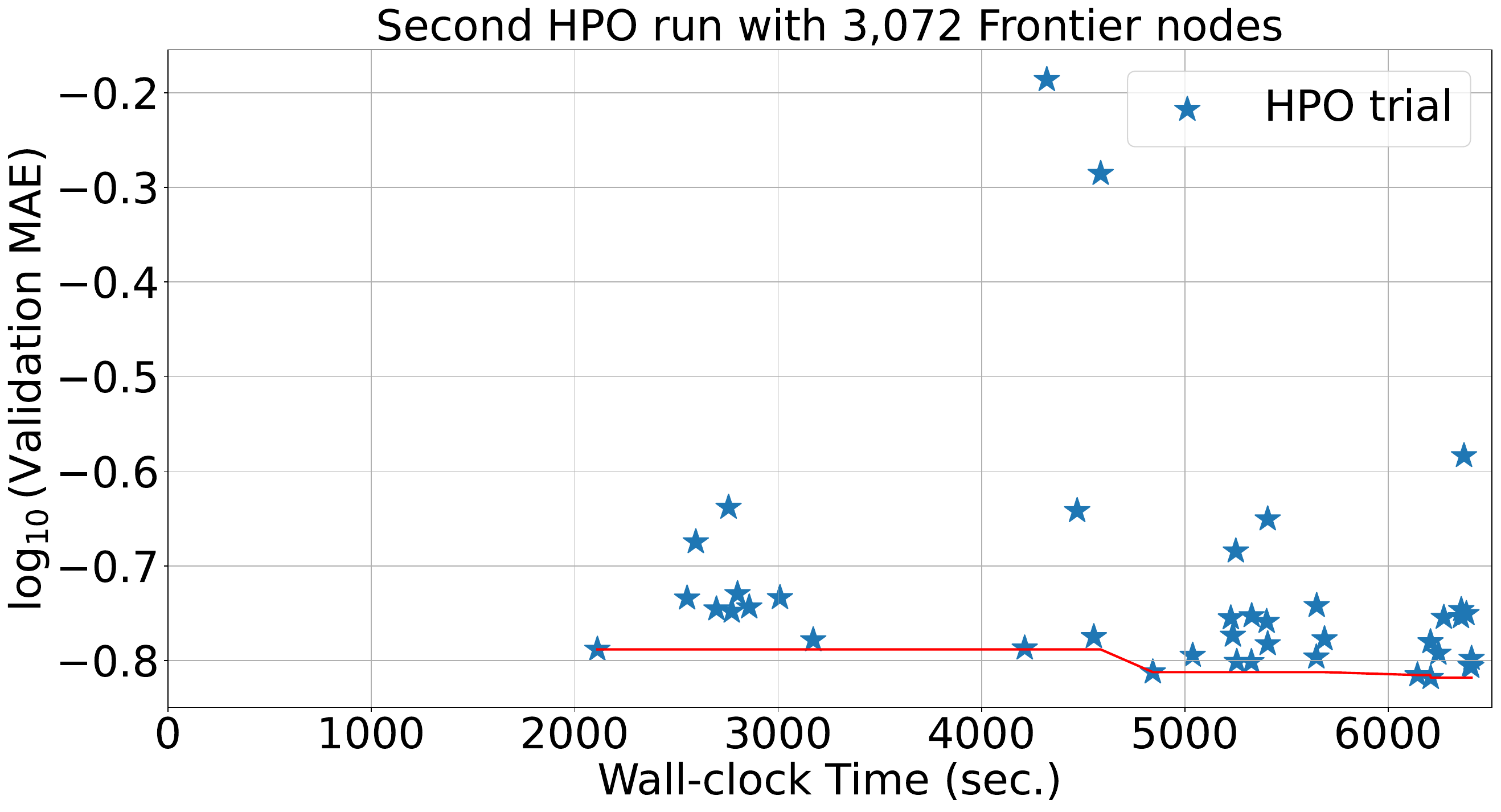}
        \caption{Logarithm in base 10 of validation MAE for HPO trials of the second HPO run on 3,072 Frontier nodes as a function of wall-clock time expressed in seconds. The red line shows the cumulative minimum across wall-clock time. }
        \label{fig:hpo_objective_2}
\end{figure}

\begin{figure}[ht!]
    \centering
        \includegraphics[width=0.6\textwidth]{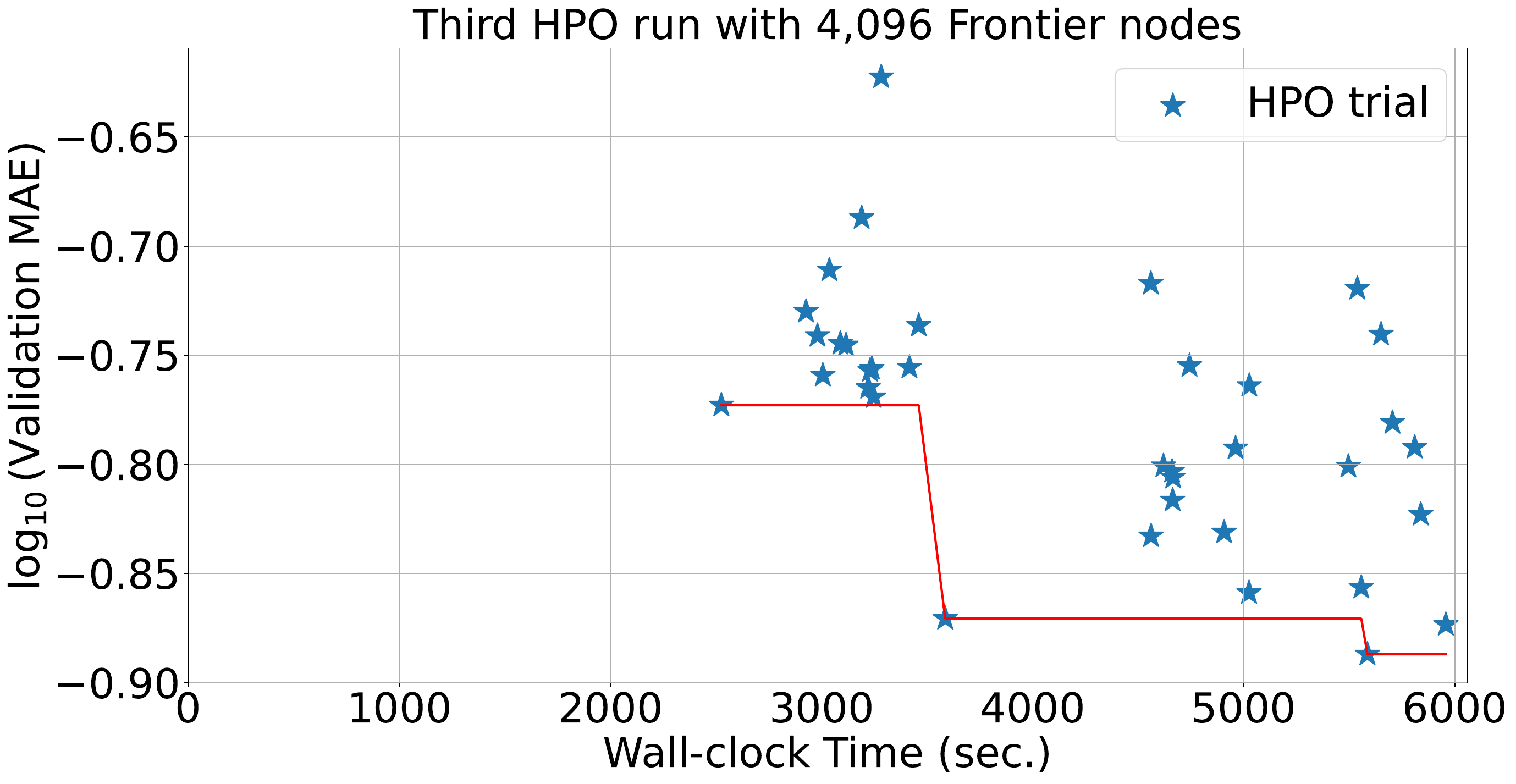}
        \caption{Logarithm in base 10 of validation MAE for HPO trials of the third HPO run on 4,096 Frontier nodes as a function of wall-clock time expressed in seconds. The red line shows the cumulative minimum across wall-clock time. }
        \label{fig:hpo_objective_3}
\end{figure}

\begin{figure}[ht!]
        \centering
        \includegraphics[width=0.6\textwidth]{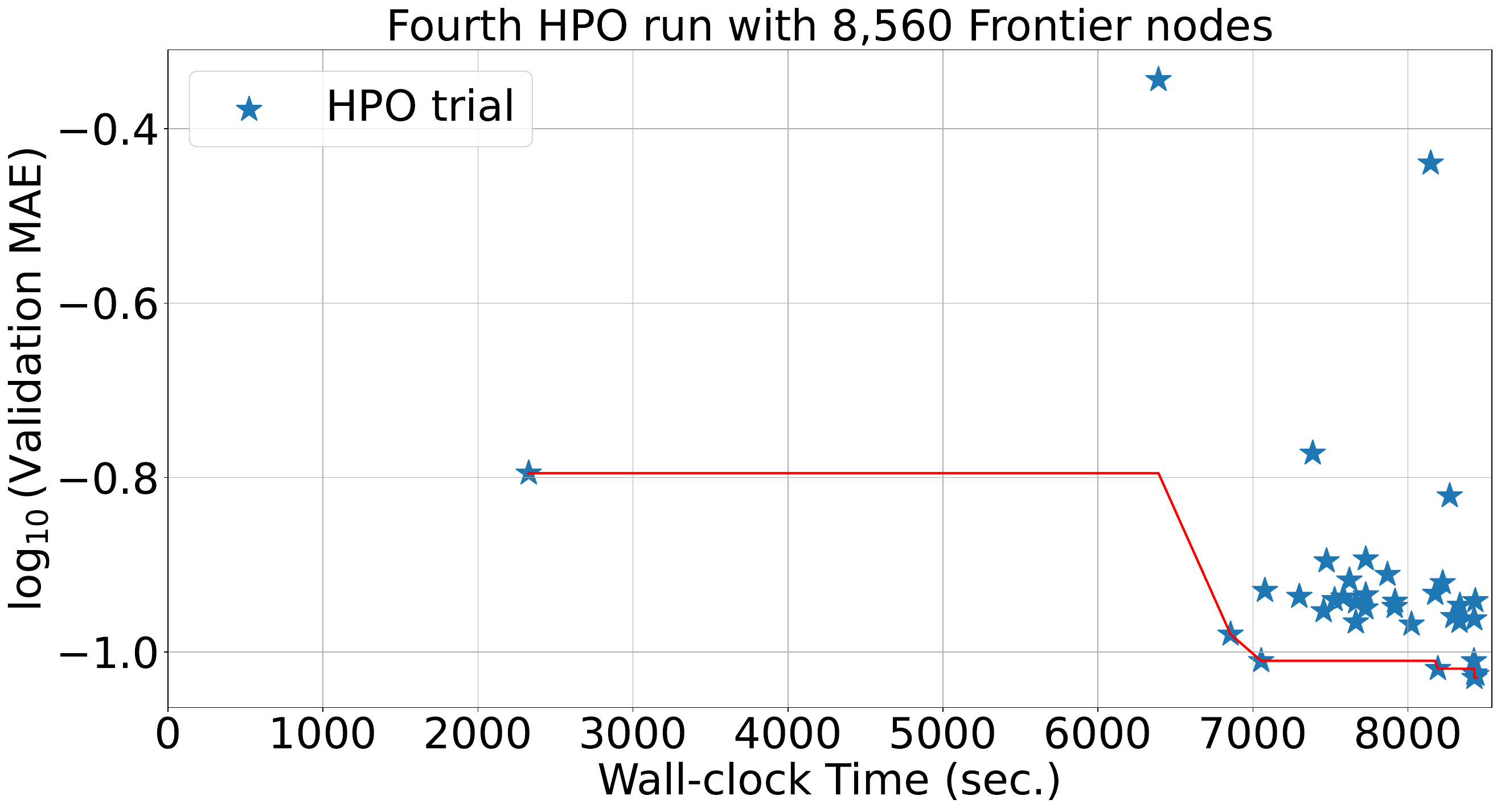}
        \caption{Logarithm in base 10 of validation MAE for HPO trials of the fourth HPO run on 8,560 Frontier nodes as a function of wall-clock time expressed in seconds. The red line shows the cumulative minimum across wall-clock time. }
        \label{fig:hpo_objective_4}
\end{figure}

Figures \ref{fig:hpo_objective_1}, \ref{fig:hpo_objective_2}, \ref{fig:hpo_objective_3}, and \ref{fig:hpo_objective_4} show the validation mean absolute error (MAE) as a function of wall-clock time for each one of the four HPO runs. For each one of the four HPO runs, the scattered distribution of blue dots (corresponding to values of the validation MAE for different HPO trials) shows that the HPO maintains a good degree of exploration throughout the entire execution. 
The red solid line indicates the minimum validation MAE obtained at a given time during the HPO run. The fact that the red line progressively descends as time progresses confirms that HPO progressively identifies GFM architectures with better accuracy. Moreover, we notice that the HPO trials of the last run are highly concentrated around the minimum (red line) across consecutive HPO runs, thereby confirming that previous HPO runs effectively provided meaningful information to hone in narrow regions of the hyperparameter space where accurate HydraGNN models can be found.


\begin{figure}[ht!]
    \centering
    \includegraphics[width=0.99\textwidth]{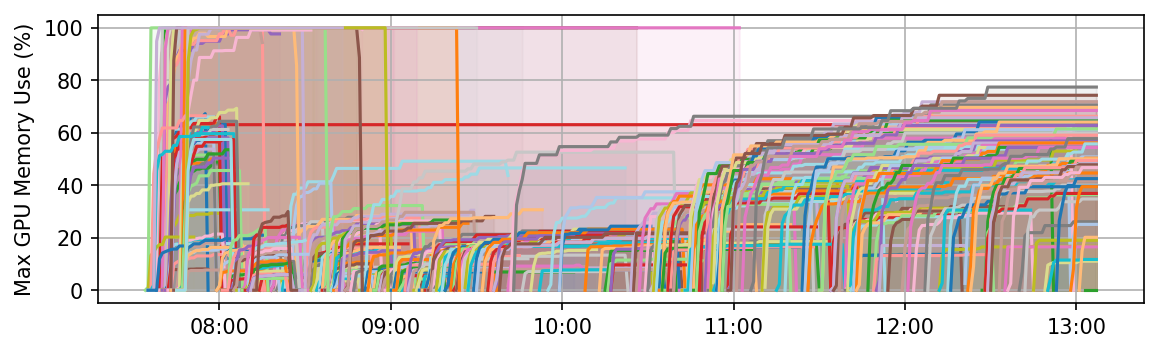}
    \caption{Max GPU HBM memory consumption traces sampled via Omnistat telemetry harness during final HPO exercise using 8,560 Frontier nodes (68,480 GCDs) executed on OLCF-Frontier.}
\label{fig:hpo_memusage}
\end{figure}

To characterize the dynamic resource behavior of HPO as multiple trials with varying model configurations are running in parallel, telemetry data collected by Omnistat was used to track the GPU memory usage for each trial as a function of time.  Fig.~\ref{fig:hpo_memusage} highlights these memory traces for all trials initiated during the final HPO exercise using 8,560 Frontier nodes during a 6-hour run. Each line on the plot corresponds to one of 390 trials initiated and we observe a dynamic high water mark peaking at 99.9\% of available memory.
The variability of memory utilization across different trials is due to the fact that different groups of GPUs (associated with different HPO trials) train HydraGNN models of different sizes, which affects the amount of GPU memory engaged at different stages of the model training. 

\subsection{Energy profiling}

\begin{figure}[ht!]
    \centering
    \includegraphics[width=0.8\textwidth]{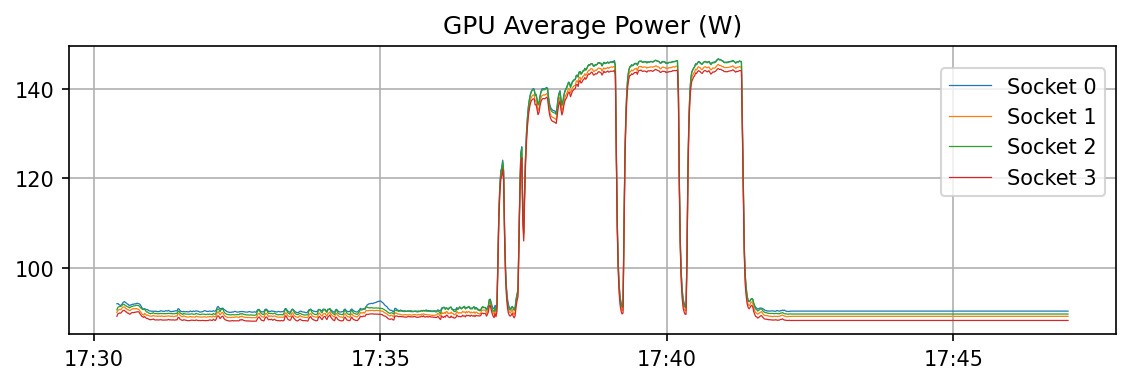}\\
    \includegraphics[width=0.8\textwidth]{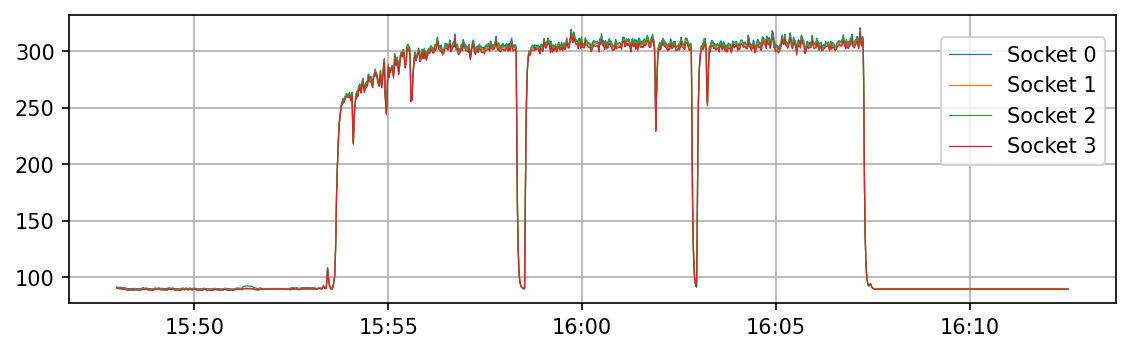}\\
    \includegraphics[width=0.8\textwidth]{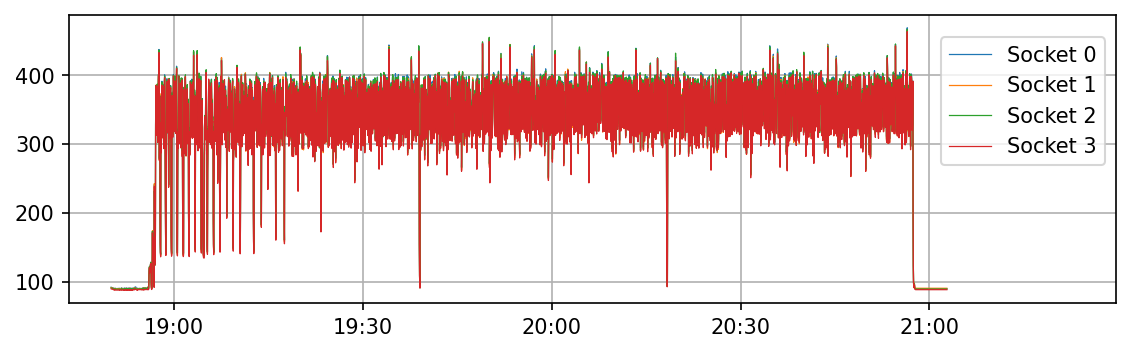}
    \caption{GPU Energy use over time for three models -- SMALL (top), MEDIUM (middle), and LARGE (bottom).  Each line represents one AMD Instinct MI250x.}
\label{fig:energy}
\end{figure}

To quantify the energy usage as a function of different model sizes, three training epochs of the SMALL, MEDIUM, and LARGE model configurations listed in Table~\ref{tab:modelsize} were completed with the Omnistat telemetry tool sampling at one second intervals. Measurements consider the entire run duration including I/O for the initial data loading process. Each model executed using 1,024~GPUs on 128~compute nodes which is the minimum node count needed to accommodate memory requirements for the LARGE model configuration. The resulting GPU energy measurements as a function of model size are summarized in Table~\ref{tab:energy}. While total energy scales with the execution time, note that GPU utilization (occupancy) also influences the energy consumed. Table~\ref{tab:energy} includes mean utilization observed across all 1,024 GPUs. The LARGE case showed the highest GPU utilization--around 89\%. The underling power histories used to compute total energy consumed for each model configuration are shown in Fig.~\ref{fig:energy}.  From these plots, we see evidence of the underlying training process with three epoch cycles visible in the power response. Furthermore, the increased GPU utilization for the larger models leads to increased GPU power demand with the LARGE model encountering peak power measurements in excess of 520 watts (W) (the peak TDP power for the AMD MI250 socket is 560W).

\begin{table}[ht]
\normalsize
  \centering
  \begin{tabular}{|r|c|c|c|}
    \hline
\textbf{Model size} & \textbf{Duration} & \textbf{\thead{\normalsize Mean GPU \\ \normalsize Utilization}} & \textbf{\thead{\normalsize GPU Energy \\ \normalsize Consumed}}  \\
\hline
SMALL & 17 mins &  12.5 \% & 14.0 kWh \\
\hline
MEDIUM & 25 mins & 46.0 \% & 42.7 kWh \\
\hline
LARGE & 133 mins & 88.9 \% & 366.6 kWh \\
\hline
\end{tabular}
\vspace*{0.1in}
  \caption{Energy usage during training on OLCF-Frontier.}
  \label{tab:energy}
\end{table}


\begin{figure}[ht!]
    \centering
    \includegraphics[width=0.8\textwidth]{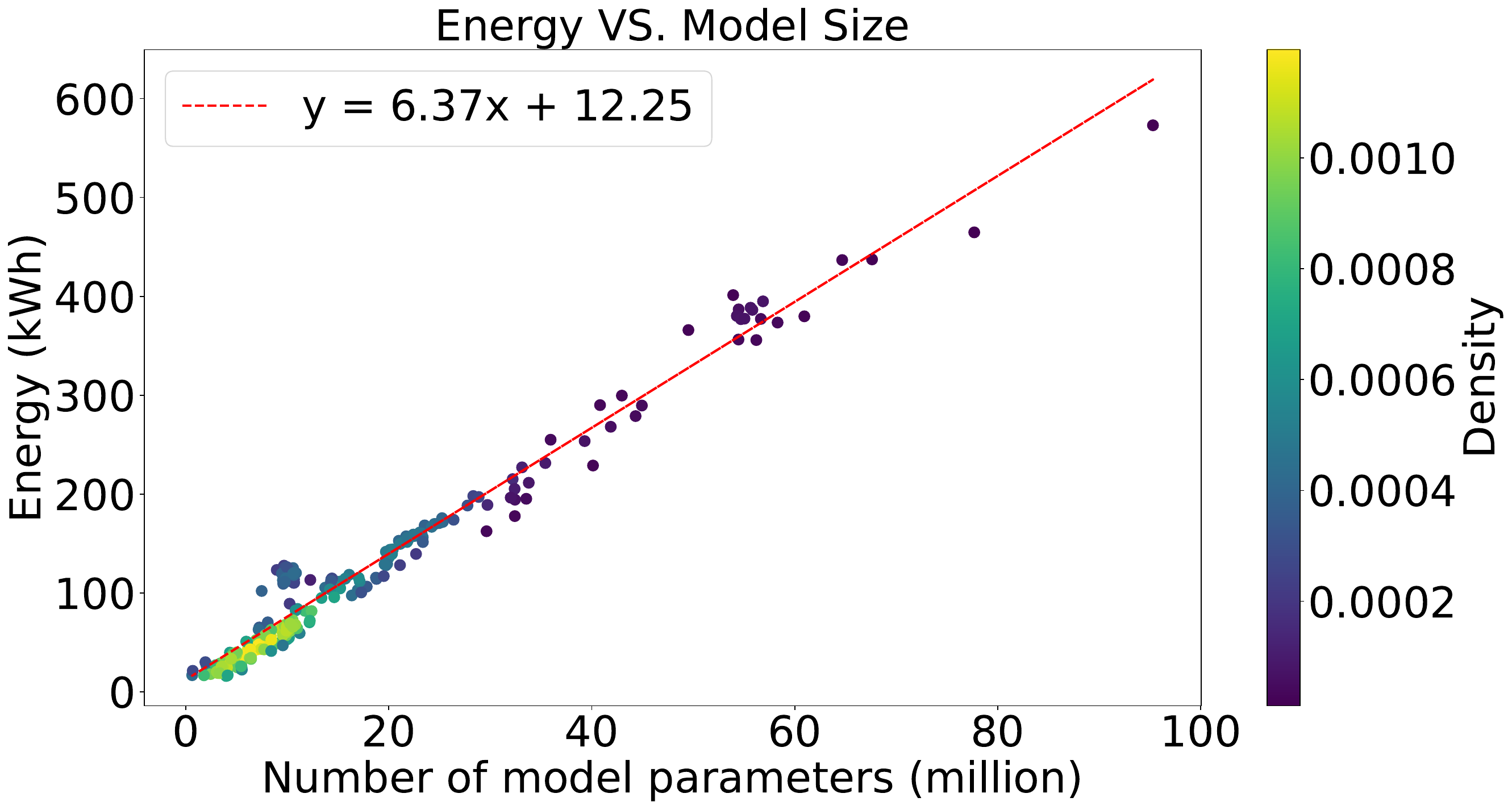}
    \caption{Energy consumption of each HPO trial of the four consecutive HPO run as a function of the number of model parameters. The red line denotes the estimated trend of a linear regression model.}
\label{fig:energy_vs_modelsize}
\end{figure}

In Figure \ref{fig:energy_vs_modelsize} we show a scatterplot of the GPU energy consumption collected using Omnistat telemetry for each HPO trial executed during the four consecutive HPO runs against the number of models in the parameters, which clearly shows a linear trend between the two quantities. 

We used this analysis to obtain energy-efficiency in the development and pre-training of GFMs as described in the following Section \ref{sec:fullpre-train}.

\subsection{Energy-efficient full training of best performing HydraGNN models identified by scalable HPO}\label{sec:fullpre-train}

\begin{figure}[ht!]
    \centering
    \includegraphics[width=0.8\textwidth]{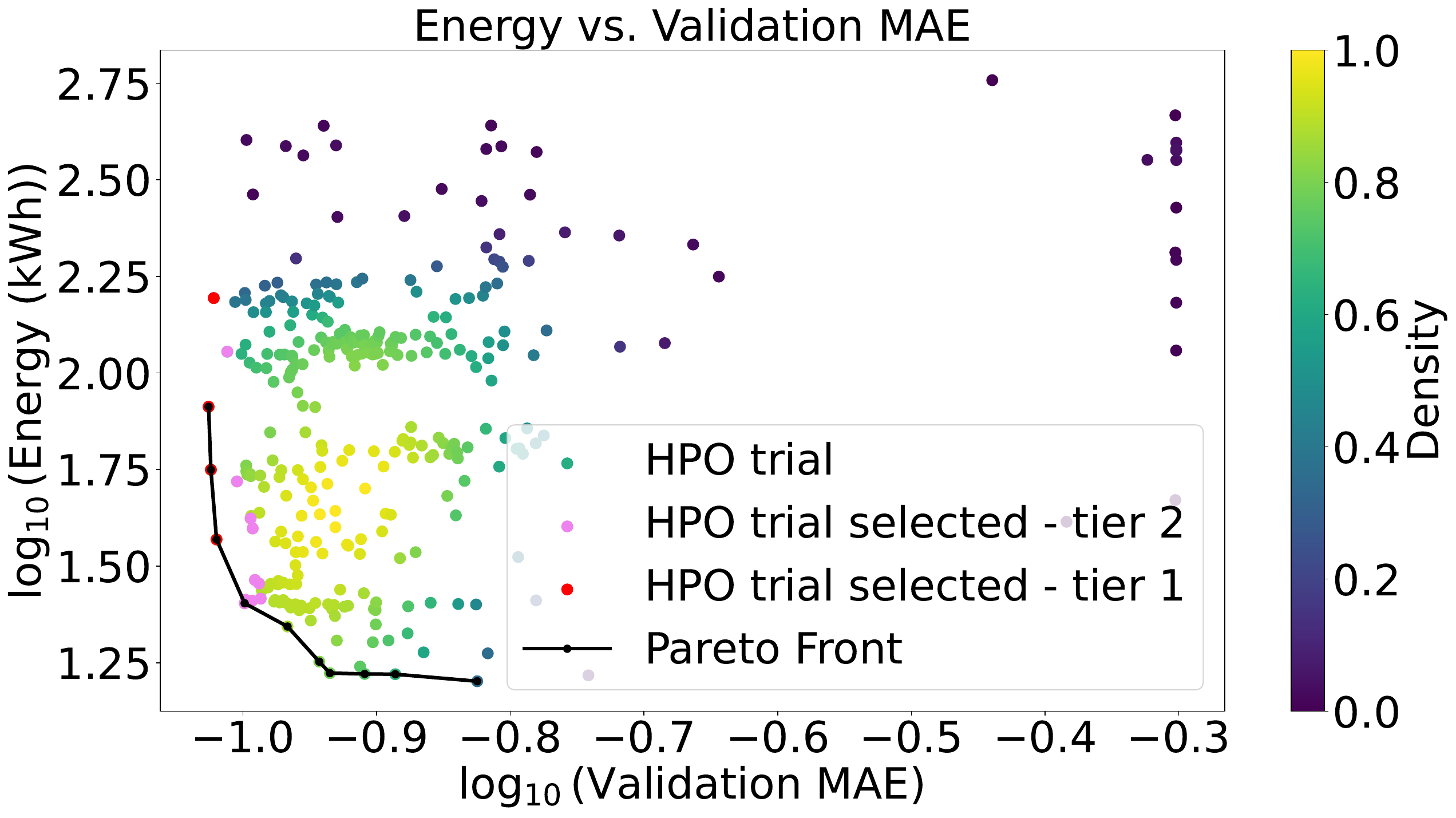}
    \caption{Energy consumption of each HPO trial plotted against the validation MAE. The black dashed line denotes the Pareto front and the red dots represented the HPO trials selected for ensemble UQ. The HPO trials of the first tier are shown in red and the HPO trials of the second tier are shown in pink.}
\label{fig:energy_vs_loss}
\end{figure}

To ensure trustworthiness of our GFM, we refined the pretraining of an ensemble of fifteen HPO trials selected from the four consecutive HPO runs and use them for the ensemble UQ. To this end, we sorted the HPO trials for increasing values of the validation MAE. The results showed that four HPO trials were outperforming all the other in terms of accuracy, with a validation MAE below 0.10. The HydraGNN architectures associated with these HPO trials have been identified as worth being further trained and thus included in the ensemble for UQ as first tier. For the selection of the remaining eleven HPO trials to be included in the ensemble, we noticed that thirty-two HPO trials were clustered within a second tier with a narrow range of the validation MAE between 0.10 and 0.125. Although these models were very close to each other in terms of accuracy, they were quite different in terms of energy consumed during the HPO run. Therefore, to ensure energy efficiency, we applied a second screening of the HPO trials based on energy consumption. To this end, we performed local sorting restricted only among these thirty-two HPO trials based on increasing amount of energy consumed. The eleven models with the lowest energy consumption have been selected and included in the ensemble for UQ as second tier. In Figure \ref{fig:energy_vs_loss} we show the scatterplot of the energy consumption against the validation MAE for each HPO trial executed during the four consecutive HPO runs. The four HPO trials belonging to the first tier (in red) sit in the left most part of the plot with lowest validation MAE, whereas the eleven HPO trials of the second tier (in pink) sit along the Pareto front of energy consumption vs. validation MAE. The plot also clearly shows that all the HPO trials selected (first and second tier) sit on the side of the Pareto front that partially favors preserving optimal accuracy over energy efficiency. 

All fifteen selected HPO trials selected use the EGNN as MPNN layer. Among the different types of MPNN layers tested, the EGNN is an equivariant model. Since equivariant models are supposed to be more data-efficient than non-equivariant models by taking full advantage of symmetries to achieve the maximal accuracy with the minimal requirement on the data, the fact that the automated HPO favors EGNN seems reasonable. 

\begin{figure}[ht!]
    \centering
    \includegraphics[width=0.6\textwidth]{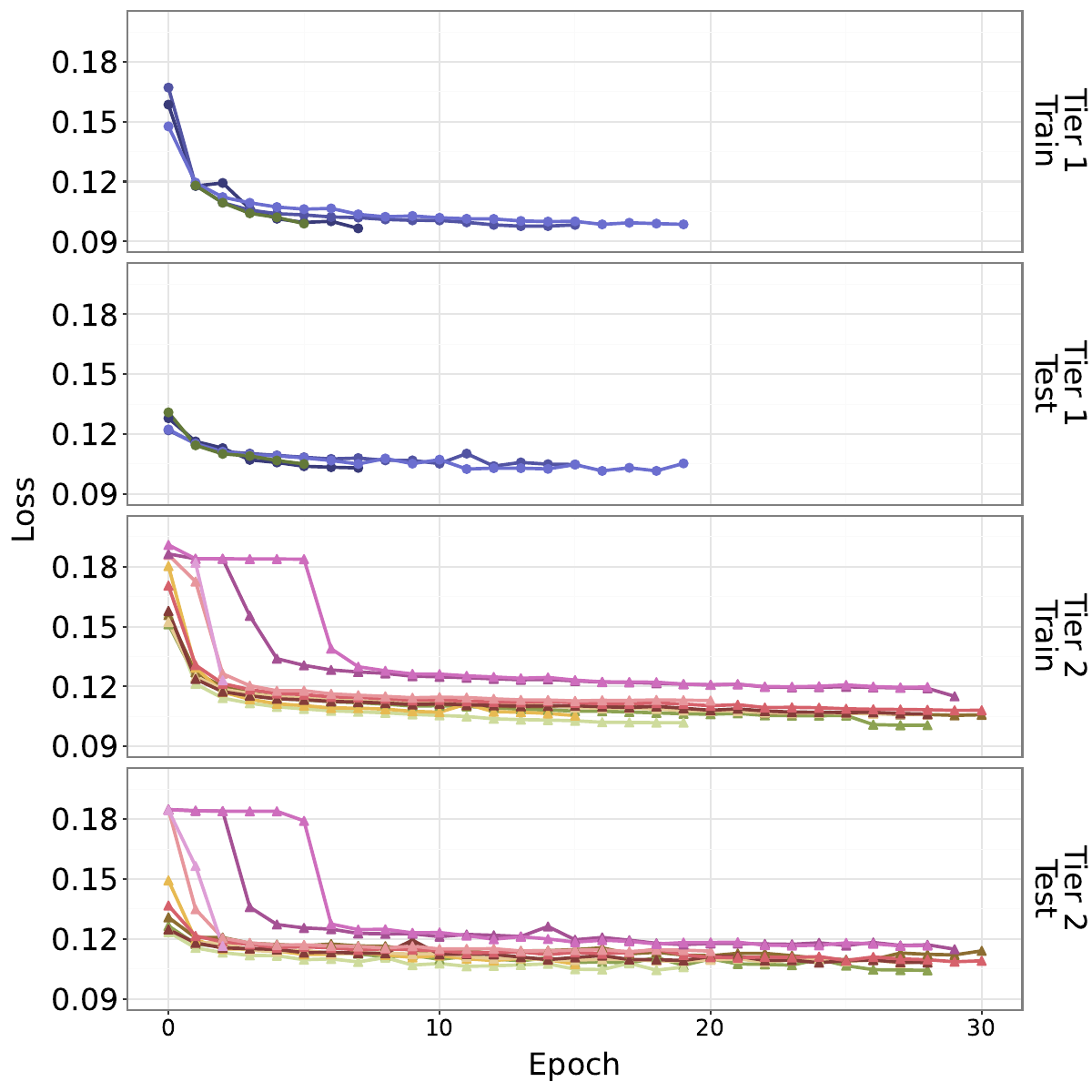}
    \caption{Full training of fifteen selected models from HPO, displaying training and test losses. Four models are shown for the first tier and eleven models are shown for the second tier.}
\label{fig:full}
\end{figure}

The fifteen selected HydraGNN models have been fully trained for a maximum of 30 epochs to reach convergence of the training, with the option of early stopping if validation MAE does not decrease across ten consecutive epochs and with the option of saving the last performed epochs if the wall-clock allocation time is about to expire. We report the trend of the training loss for all fifteen models in Fig. \ref{fig:full}. The training loss flattens at the end of the training history, indicating that the models have reached their maximum predictive capacity. Moreover, the models seem to reach similar final accuracy, thereby confirming that the HPO has indeed thoroughly spanned the hyperparameter space and identified HydraGNN models with similar predictive performance. 
We then used the ensemble of the fully pre-trained fifteen GFMs to provide ensemble predictions with epistemic uncertainty measurement, as described in the following Section \ref{sec:UQ}. 

\subsection{Ensemble predictions and epistemic UQ on ensemble of pre-trained GFMs}\label{sec:UQ}

\begin{figure}[ht!]
    \centering
    \includegraphics[width=0.8\textwidth]{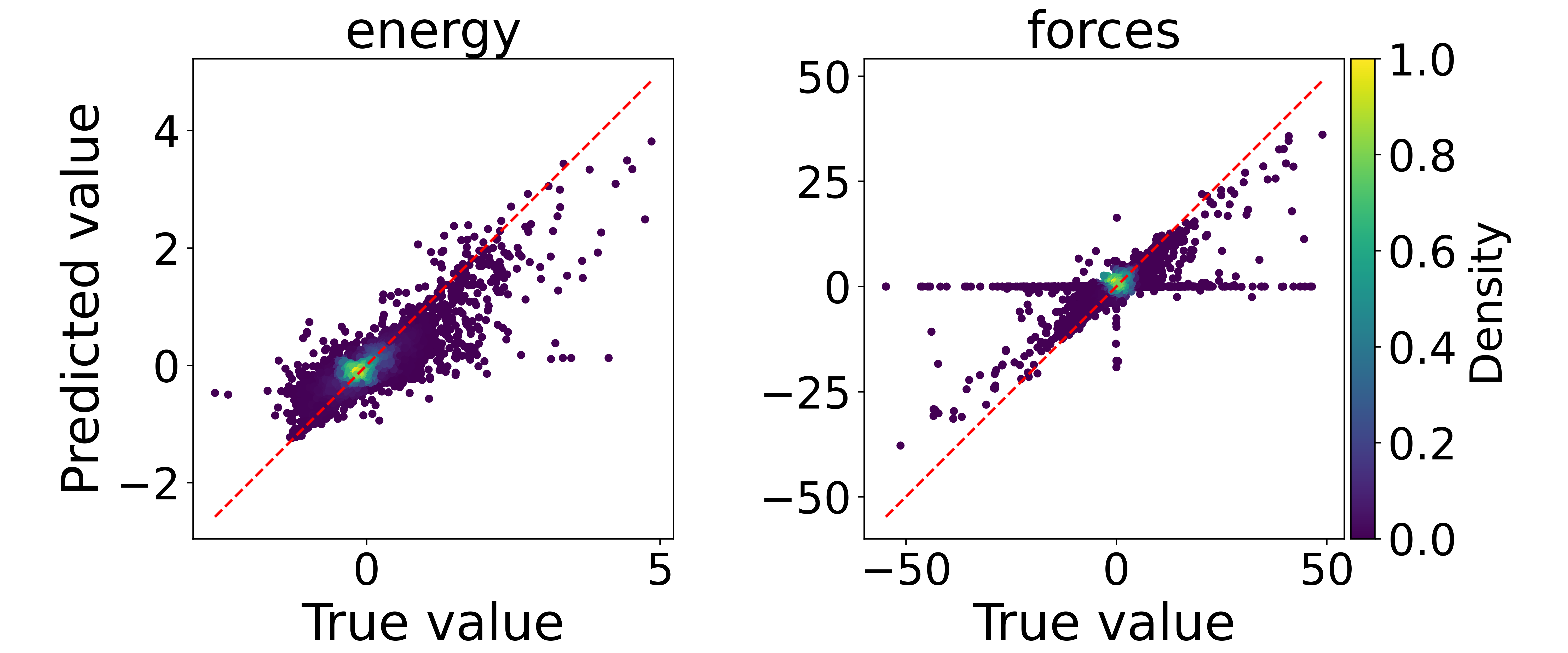}
    \caption{Parity plot of ensemble predictions (using fifteen pretrained models from Section \ref{sec:fullpre-train}) of energy and atomic forces for 20,480 atomistic structures from the test split in pretraining. The red diagonal line represents the perfect model prediction. We present the parity plots for each dataset separately in the supplementary material.}
\label{fig:ensem-pred}
\end{figure}

UQ is essential for the trustworthiness of GFMs. 
We calculate the uncertainties of the ensemble predictions in pretraining using the fifteen models from Section \ref{sec:fullpre-train}. Table \ref{tab:metrics} provides the MAE and the root mean-squared error (RMSE) of the ensemble predictions of energy and forces on the training-validation-testing splits of each dataset. Given the limited number of training epochs used for the pre-training of the GFM ensemble, the accuracy achieved is very promising and indicative that the GFMs are indeed learning the underlying chemical principles that describe interactions between atoms of different constituents. 
Fig. \ref{fig:ensem-pred} presents the parity plot of ensemble averaged predictions for 20,480 atomistic structures from the test split, showing that our ensemble models have learned the generic information from the diverse pretraining data. While the accuracy is not as high as that reported in task-specific AI work, our pre-trained GFMs are expected to stabilize zero-shot predictions via ensemble averaging and reduce the fine-tuning efforts across a broad set of domain-specific tasks. 

\begin{table}[h!]
\centering
\begin{tabular}{|c|c|c|c|c|c|}
\hline
\textbf{Dataset} & \textbf{Split} & \makecell{\textbf{Energy (MAE)} \\ \textbf{eV/atom}} & \makecell{\textbf{Energy (RMSE)} \\ \textbf{eV/atom} } & \makecell{ \textbf{Forces (MAE)} \\ \textbf{eV/angstrom} }& \makecell{\textbf{Forces (RMSE)} \\ \textbf{eV/angstrom}}\\ \hline
\multirow{3}{*}{ANI1x} & Train & 0.001742 & 0.003885 & 0.012941 & 0.034096 \\
                          & Validation & 0.001640 & 0.002754 & 0.012350 & 0.027084 \\
                          & Test & 0.001660 & 0.003009 & 0.012782 & 0.031976 \\ \hline
\multirow{3}{*}{MPTrj} & Train & 0.260078 & 0.407861 & 0.172559 & 0.848933 \\
                          & Validation & 0.255583 & 0.389681 & 0.149686 & 0.831007 \\
                          & Test & 0.248065 & 0.376361 & 0.146531 & 0.638788 \\ \hline
\multirow{3}{*}{OC2020} & Train & 0.017441 & 0.025182 & 0.095165 & 0.183093 \\
                                & Validation & 0.017758 & 0.026273 & 0.094872 & 0.185488 \\
                                & Test & 0.021786 & 0.032004 & 0.106405 & 0.219948 \\ \hline
\multirow{3}{*}{OC2022} & Train & 0.045176 & 0.378741 & 0.082496 & 0.334222 \\
                           & Validation & 0.049585 & 0.079223 & 0.079433 & 0.312630 \\
                           & Test & 0.065060 & 0.144687 & 0.082543 & 0.314317 \\ \hline
\multirow{3}{*}{qm7x} & Train & 0.013291 & 0.020044 & 0.141055 & 0.226754 \\
                         & Validation & 0.013056 & 0.020049 & 0.140915 & 0.233489 \\
                         & Test & 0.013865 & 0.021830 & 0.143753 & 0.235468 \\ \hline
\end{tabular}
\caption{MAE and RMSE values for energy and forces across datasets and splits.}
\label{tab:metrics}
\end{table}

\begin{figure}[ht!]
    \centering
    \includegraphics[width=0.8\textwidth]{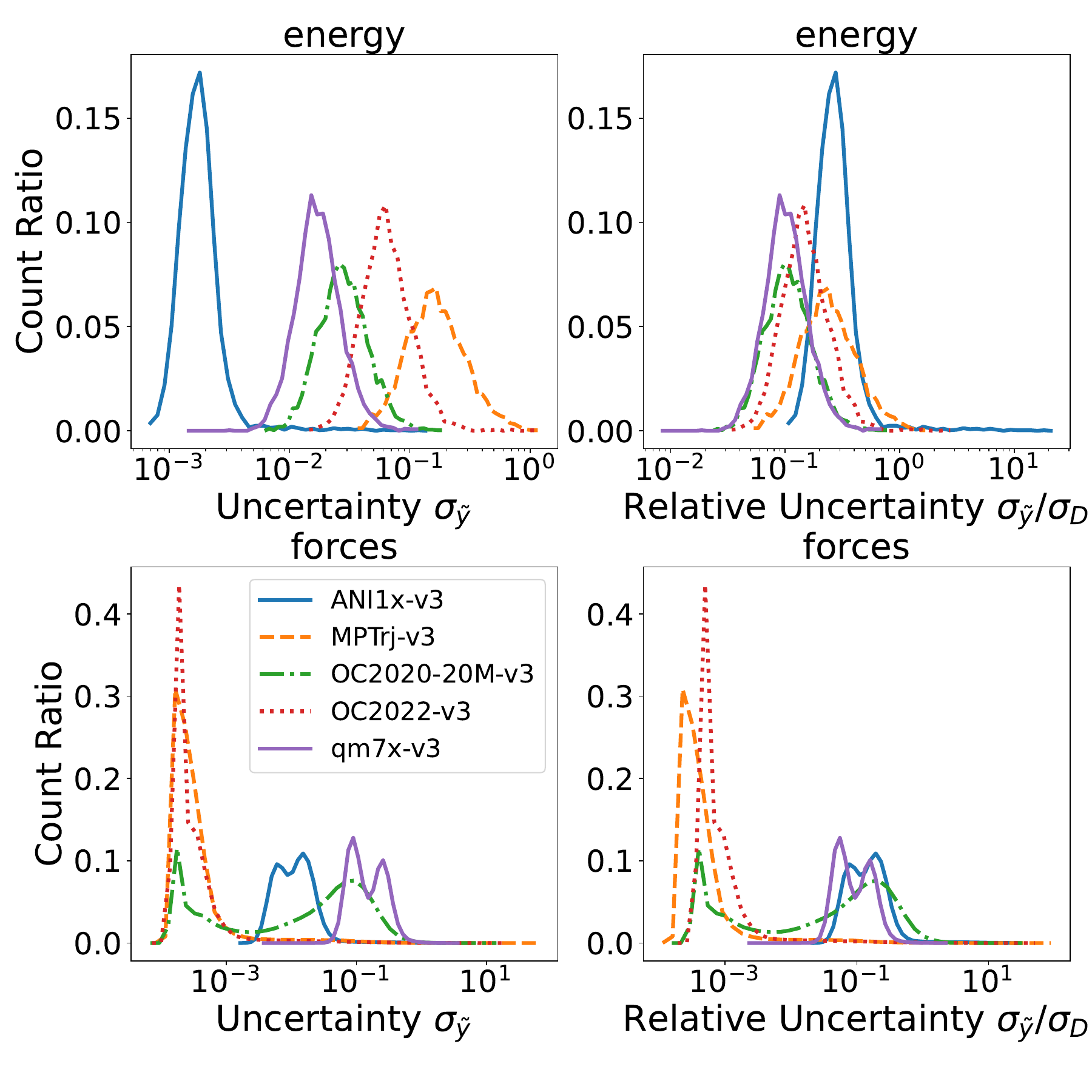}
    \caption{Histogram of epistemic uncertainty $\sigma_{\Tilde{y}}$ (left column) and relative uncertainty $\sigma_{\Tilde{y}}/\sigma_{D}$ (right column) (with $\sigma_{D}$ being the data STD for each dataset and being summarized in the supplementary material) in energy and force predictions for each dataset used in pretraining.}
\label{fig:uncertainty}
\end{figure}

Figure \ref{fig:uncertainty} shows the histogram of epistemic uncertainty (left column) of the corresponding energy and force predictions in pretraining, calculated on 4,096 atomistic structure from the test split of each dataset (in total, 20,480 structures). 
The uncertainty $\sigma_{\Tilde{y}}$ is defined as the STD of ensemble predictions in Equation \eqref{eq-enstd}, which is commonly used but biased toward large true values and may not accurately reflect actual prediction quality.
We also plot the relative uncertainty to data STD (right column) in Figure \ref{fig:uncertainty}. 
Three datasets are observed to higher relative uncertainties---\texttt{ANI1x}, \texttt{qm7x}, and \texttt{MPTrj}---due to two factors. 
First, datasets \texttt{ANI1x} and \texttt{qm7x} each contain approximately 4-5 million organic compounds, in contrast to the other datasets, which include more than 145 million inorganic compounds. Second, dataset \texttt{MPTrj} includes higher fidelity solutions but is much smaller in size compared to the others.
Some of the atomistic structures from the datasets \texttt{qm7x} and \texttt{ANI1x} are associated with significantly inaccurate predictions of atomic forces, which correspond to the data samples sitting on the horizontal point cloud in the parity plot for the prediction of atomic forces. Since the ensemble of GFMs correctly associate high uncertainty to atomistic structures whose predictions of atomic forces are notably inaccurate, this corroborates the fact that our GFMs are self-aware of their own regime of trustworthiness, which is an important indicator of robustness.   

This uncertainty information---varying uncertainty values stemming from imbalanced representation in training data---not only provides a confidence measure of our GFM predictions but also guide data collection for further model development in an active learning setting.

%

\section{Conclusions and Future Work}
\label{sec:conclusion}


In this work we described our approach towards developing and training trustworthy and energy-efficient predictive GFMs by scaling the HydraGNN architecture on over 154 millions of atomistic materials modeling data using two DOE leadership class supercomputers, viz. NERSC-Perlmutter and OLCF-Frontier. We discussed optimizations and tools used for developing a GFM and running HPO at large scale.

We used distributed data management capabilities to partition large volumes of data across distributed computing resources and efficiently exchange data samples across devices using low-latency communication methods. This helped preserve global data shuffling, which is crucial for maintaining good convergence of the GFM training. 
By scaling HPO on over 91\% of the exascale OLCF-Frontier supercomputer, we have assessed the importance of thoroughly exploring a large set of hyperparameter configurations to identify HydraGNN architectures with high predictive accuracy. The use of Omnistat tools at extreme scale allowed us to conduct a thorough analysis to select HydraGNN architectures with high predictive accuracy and low energy cost.
Moreover, access to exceptionally performing large scale computing facilities allowed us to develop and test ensemble UQ capabilities to measure the degree of confidence associated with the HydraGNN predictions, which contributes to achieving trustworthiness.  
Performing HPO and ensemble UQ at unprecedented scale on supercomputing facilities confirms our computational readiness in using HydraGNN to develop trustworthy and energy-efficient GFMs to support the US-DOE materials science needs by providing robust and transferable computational capabilities for AI-accelerated materials discovery and design. 

Future work will be devoted to improving the scaling performance of the GFM pre-training on thousands of GPUs by exploring and developing more sophisticated strategies to mitigate load imbalance between GPUs, while maintaining high quality in the attainable trustworthiness of the pre-trained GFMs. We also plan to apply techniques to read the metadata in parallel to improve the overall bandwidth of reading large-volume of datasets on supercomputers. Once the GFMs are pre-trained, we will deploy them to downstream tasks for fine-tuning, where we will illustrate the efficacy of our GFMs in reducing the amount of training data and computational resources needed to develop robust and transferable DL models for domain-specific applications.

\section*{Code availability}
The code used to develop, pre-train, and use the pre-trained GFMs is available at the following branch
\url{https://github.com/ORNL/HydraGNN/tree/Predictive_GFM_2024}
of the ORNL GitHub repository for HydraGNN. More specifically, the scripts used to pre-process the data, generate the ADIOS files, run HPO, and continue the pre-training are available in the directory called \texttt{examples/multidataset\_hpo}. The scripts that use the pre-trained ensemble of GFMs for ensemble averaging and epistemic UQ are available in the directory called \texttt{examples/ensemble\_learning}.

\section*{Data availability}
The set of ADIOS files that contain the pre-processed data and the ensemble of pre-trained GFMs that can be used for zero-shot inference and fine-tuning on downstream tasks are available open-source on the OLCF Data Constellation Facility. {\color{red} Release currently under review at OLCF with DOI 10.13139/OLCF/2474799}.


\section*{Acknowledgements}
Massimiliano Lupo Pasini would like to thank Dr. Vladimir Protopopescu for his valuable feedback in the preparation of the manuscript.
This research is sponsored by the Artificial Intelligence Initiative as part of the Laboratory Directed Research and Development (LDRD) Program of Oak Ridge National Laboratory, managed by UT-Battelle, LLC, for the US Department of Energy under contract DE-AC05-00OR22725.
This work used resources of the Oak Ridge Leadership Computing Facility, which is supported by the Office of Science of the U.S. Department of Energy, under INCITE award CPH161. This work also used resources of the National Energy Research
Scientific Computing Center, which is supported by the Office of Science of the U.S. Department of Energy under Contract No. DE-AC02-05CH11231, under award ERCAP0025216 and ERCAP0027259.

\newpage

\begin{appendices}

\section{Full training}

\begin{figure}[ht!]
    \centering
    \includegraphics[width=0.5\textwidth]{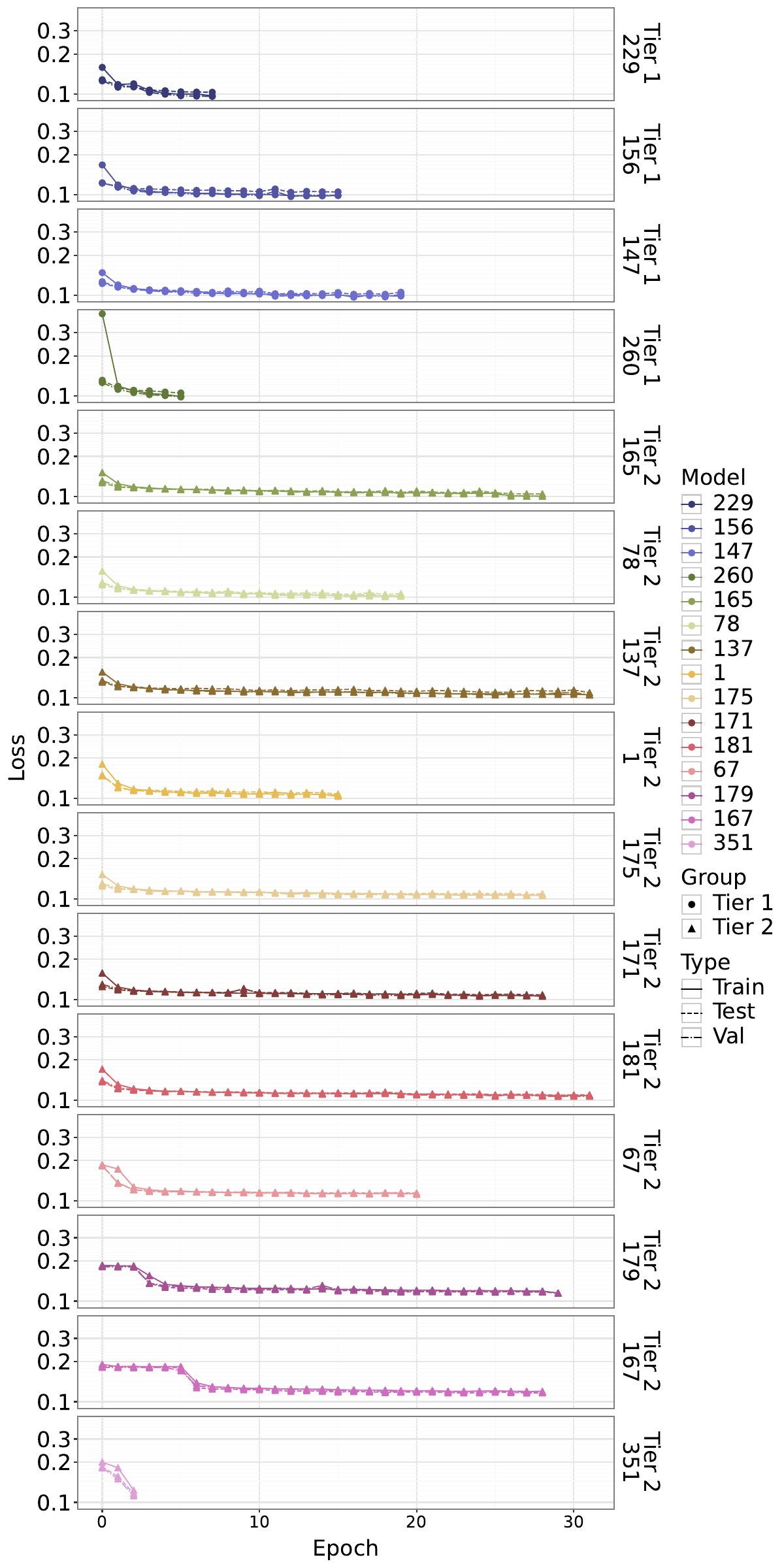}
    \caption{
    Full training of fifteen selected models from HPO trials for 30 epochs, categorized into Tier 1 and Tier 2 groups.}
\label{fig:full-all}
\end{figure}

Figure \ref{fig:full-all} presents the fifteen selected models from the HPO runs categorized in Tier 1 and Tier 2. Each model was trained for a maximum of 30 epochs. The plot shows training, validation, and test losses for each model. 
Some models exhibit early stops due to various reasons, such as no  further progress, reaching a time limit, or encountering NaN values during training for unknown causes.

\section{More results on ensemble predictions and UQ}

\begin{figure}[ht!]
    \centering
     \includegraphics[width=0.8\textwidth]{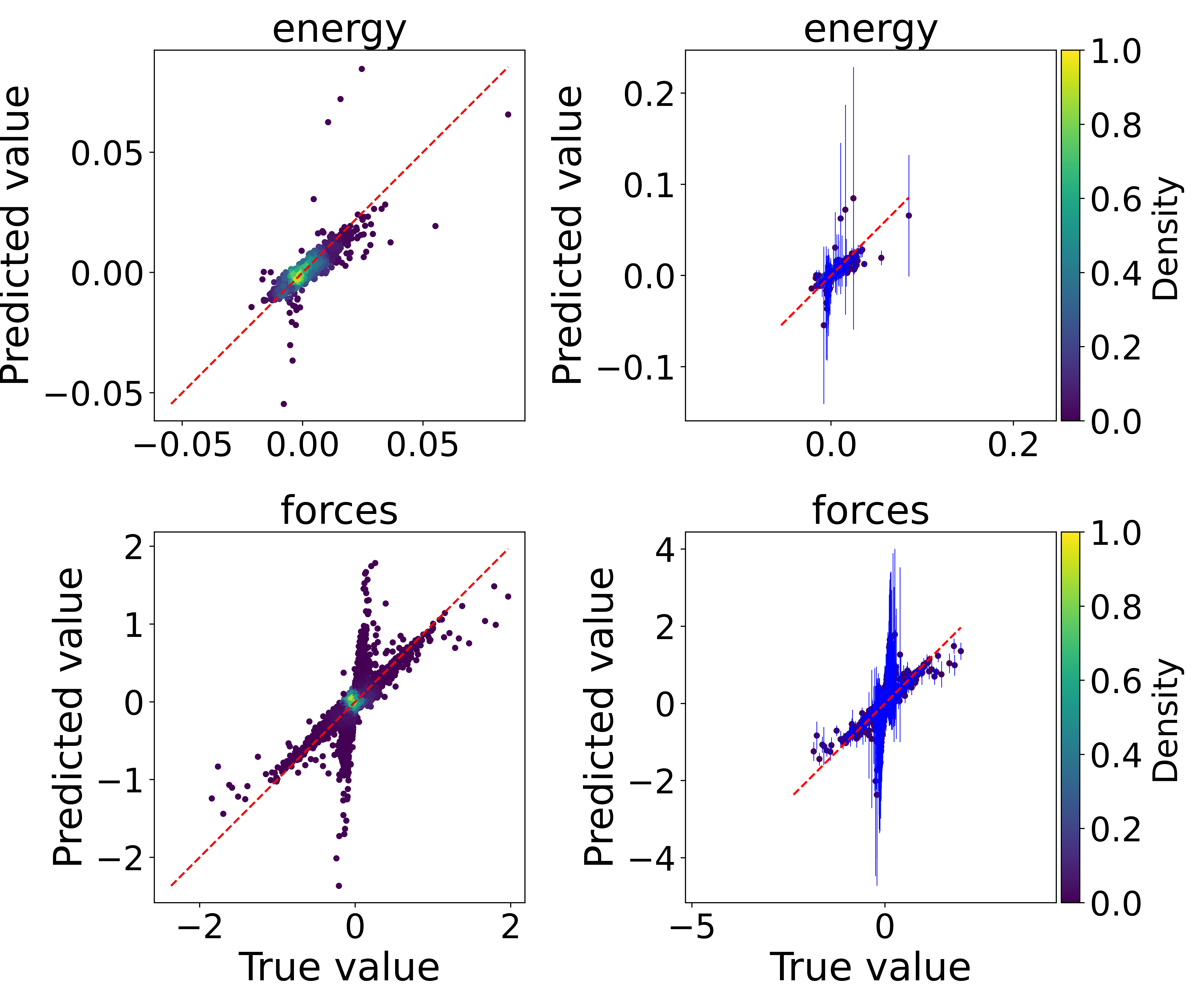}
    \caption{Parity plot of ensemble predictions of energy and atomic forces for 4,096 atomistic structures from the test split of ANI1x in pretraining. The error bars represent the predictive uncertainties. The red diagonal line represents the perfect model prediction.}
\label{fig:ensem-predani}
\end{figure}

\begin{figure}[ht!]
    \centering
     \includegraphics[width=0.8\textwidth]{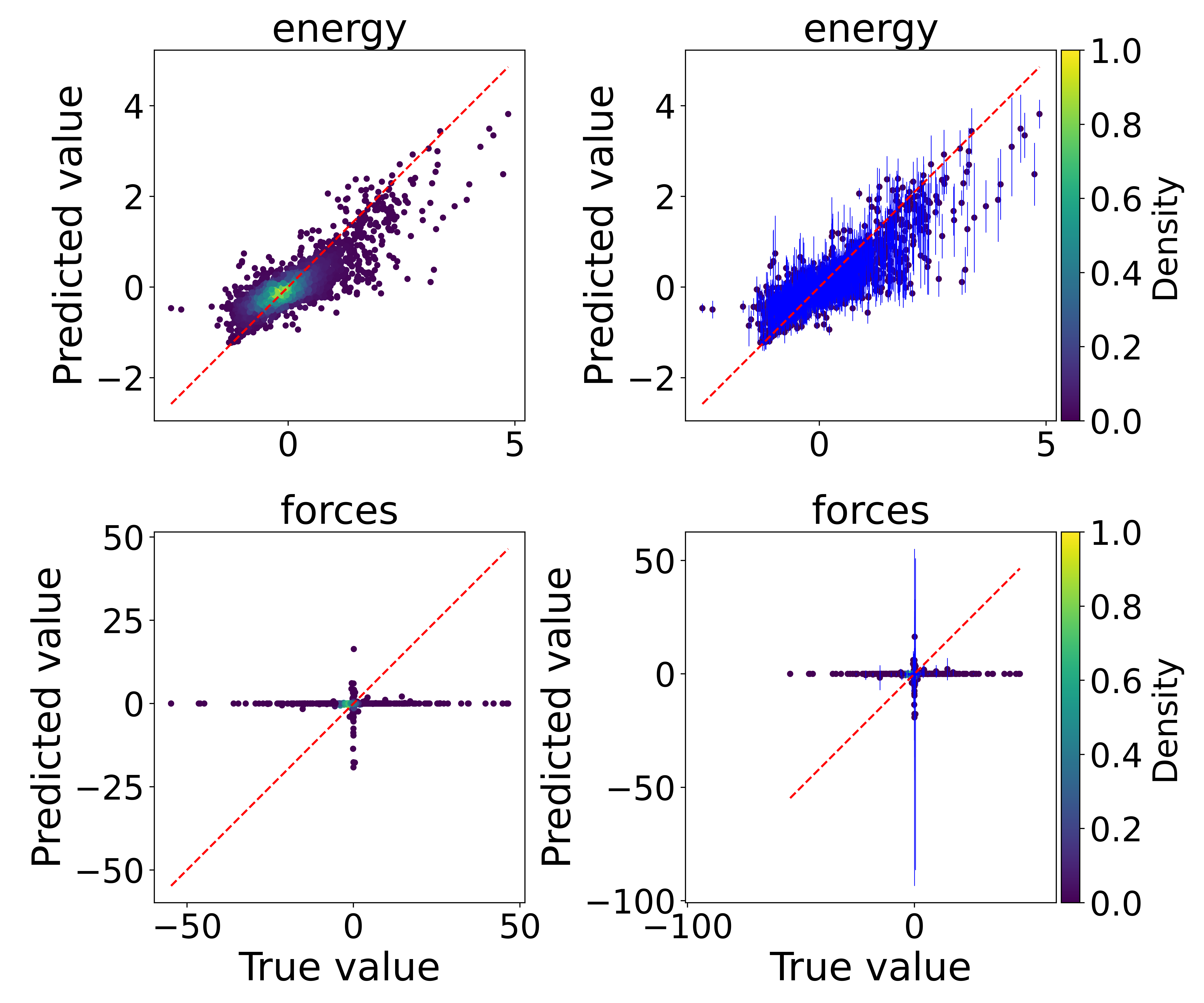}
    \caption{Parity plot of ensemble predictions of energy and atomic forces for 4,096 atomistic structures from the test split of MPTrj in pretraining. The error bars represent the predictive uncertainties. The red diagonal line represents the perfect model prediction.}
\label{fig:ensem-predmpt}
\end{figure}

\begin{figure}[ht!]
    \centering
     \includegraphics[width=0.8\textwidth]{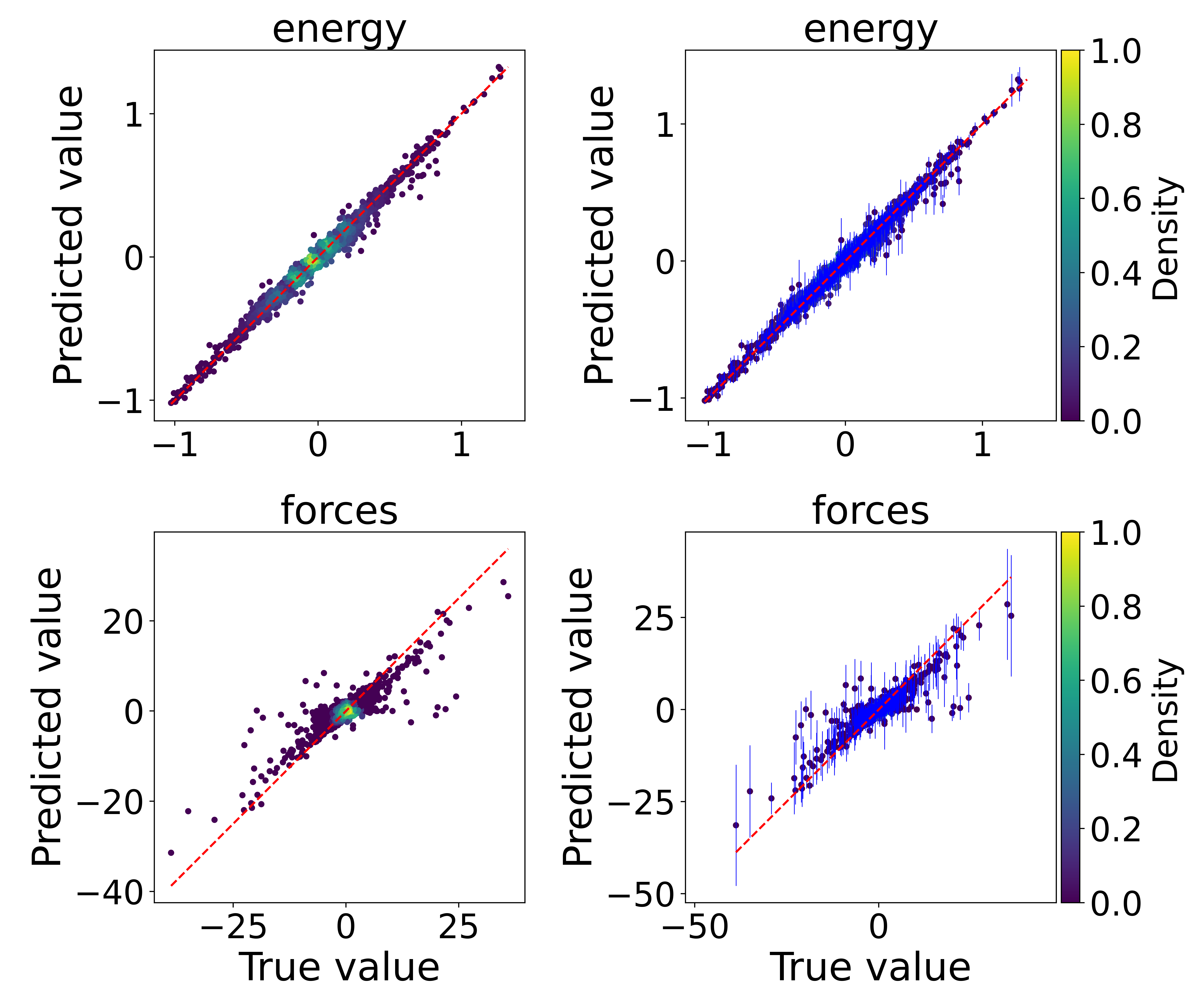}
    \caption{Parity plot of ensemble predictions of energy and atomic forces for 4,096 atomistic structures from the test split of OC2020 in pretraining. The error bars represent the predictive uncertainties. The red diagonal line represents the perfect model prediction.}
\label{fig:ensem-predoc20}
\end{figure}

\begin{figure}[ht!]
    \centering
     \includegraphics[width=0.8\textwidth]{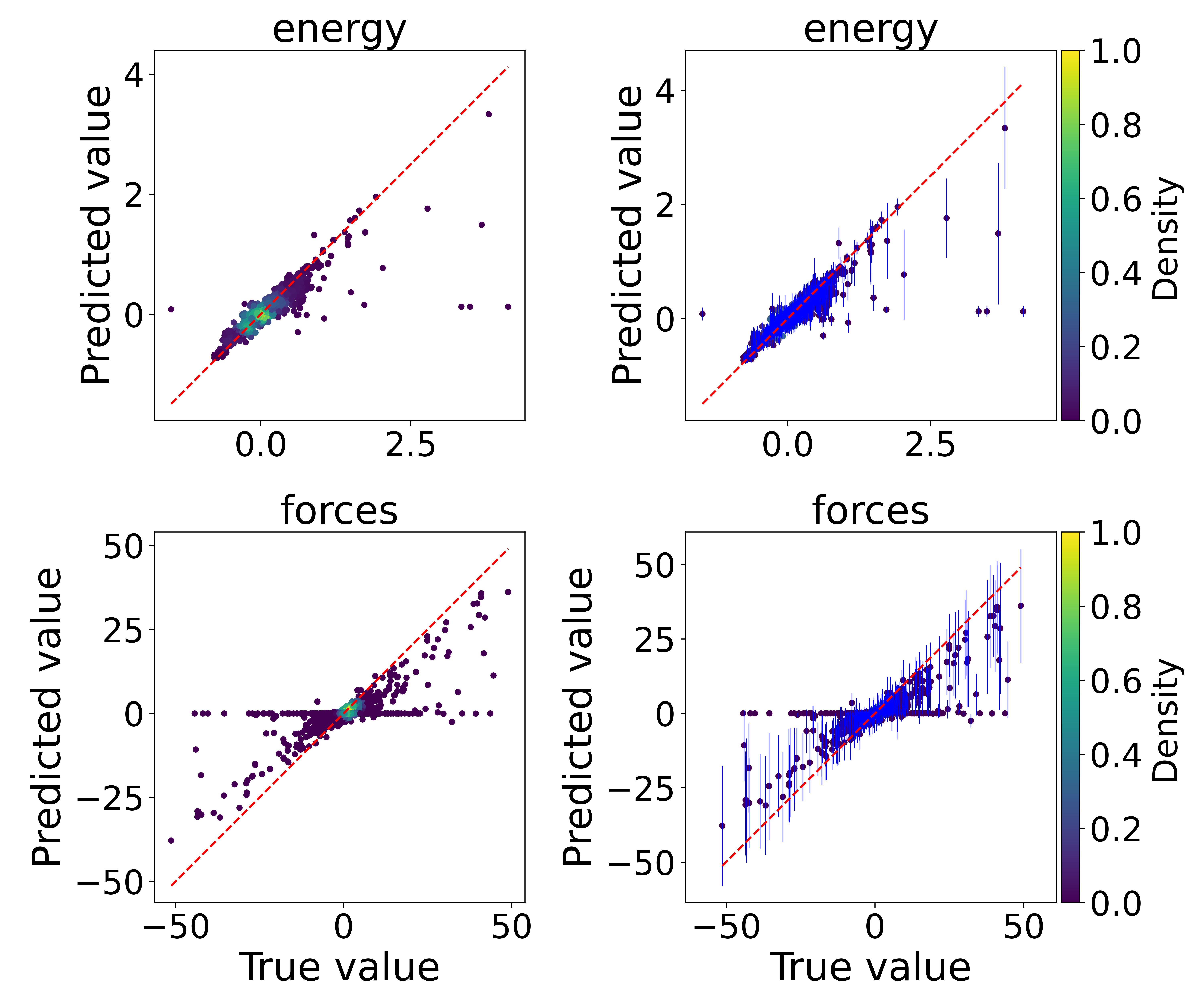}
    \caption{Parity plot of ensemble predictions of energy and atomic forces for 4,096 atomistic structures from the test split of OC2022 in pretraining. The error bars represent the predictive uncertainties. The red diagonal line represents the perfect model prediction.}
\label{fig:ensem-predoc22}
\end{figure}

\begin{figure}[ht!]
    \centering
     \includegraphics[width=0.8\textwidth]{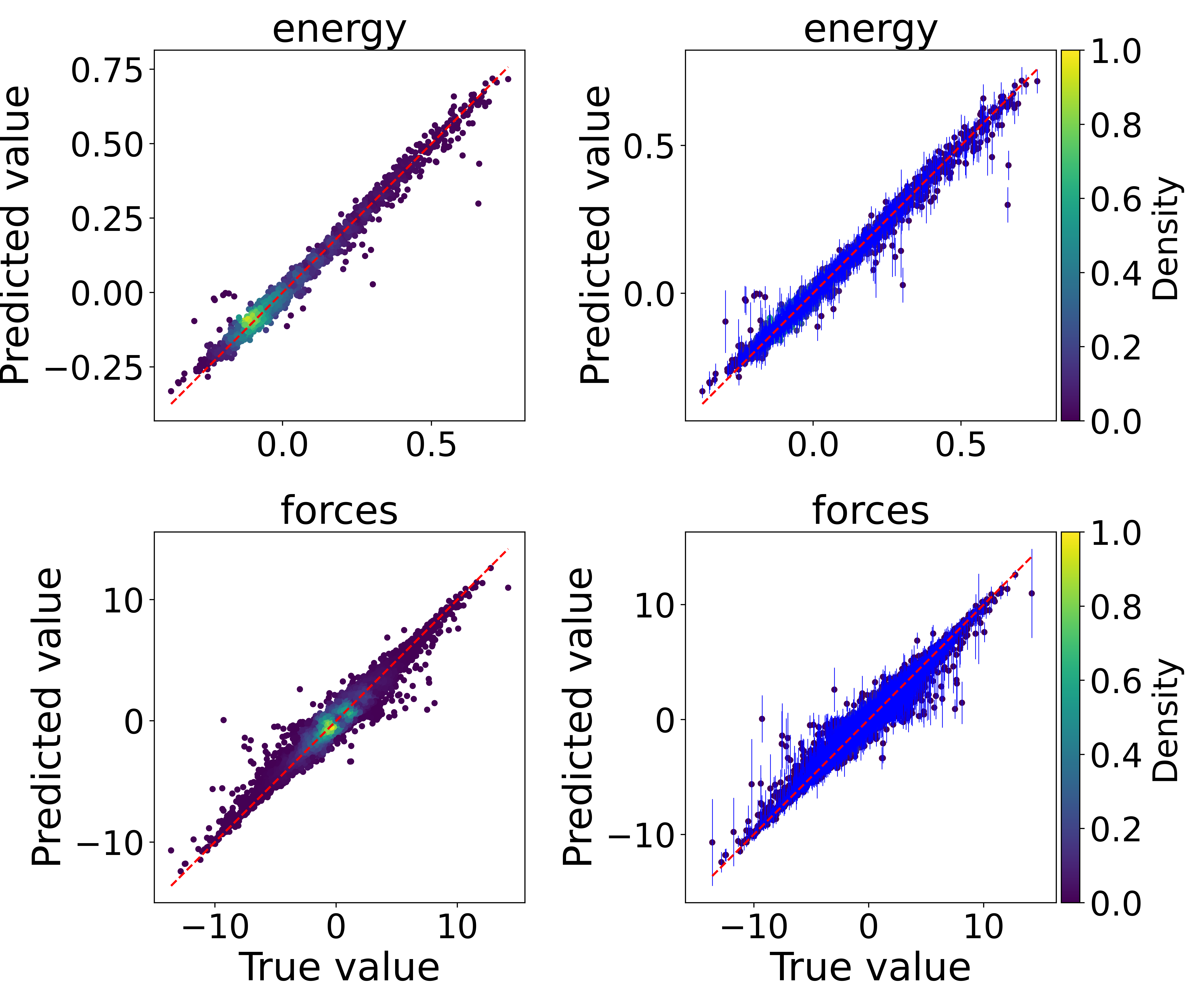}
    \caption{Parity plot of ensemble predictions of energy and atomic forces for 4,096 atomistic structures from the test split of qm7x in pretraining. The error bars represent the predictive uncertainties. The red diagonal line represents the perfect model prediction.}
\label{fig:ensem-predqm7x}
\end{figure}

Figures \cref{fig:ensem-predani,fig:ensem-predmpt,fig:ensem-predoc22,fig:ensem-predoc20,fig:ensem-predqm7x} show the parity plot of ensemble averaged predictions for 4,096 atomistic structures from the test split of the five datasets (i.e., \texttt{ANI1x}, \texttt{MPTrj}, \texttt{OC2020}, \texttt{OC2022}, and \texttt{qm7x}), respectively, showing that our ensemble models have learned the generic information from the diverse pretraining data. The models exhibit larger uncertainties in \texttt{ANI1x} and \texttt{MPTrj}.

Table \ref{tab:data-std} summarizes the standard deviation values used in relative uncertainty calculations for energy and forces in each dataset.

\begin{table}[ht!]
\normalsize
\centering
\begin{tabular}{|l|r|r|}
\hline
\textbf{Dataset} & \textbf{STD of energy} & \textbf{STD of force components} \\
\hline
\texttt{ANI1x}  & 6.48e-3 & 7.83e-2\\
\texttt{QM7-X} & 1.70e-1 & 1.62 \\
\texttt{OC2020}  & 2.64e-1 & 4.37e-1 \\ 
\texttt{OC2022} & 4.26e-1 & 3.77e-1 \\
\texttt{MPTrj} & 6.93e-1 & 7.23e-1\\
\hline
\end{tabular}
\vspace{0.1in}
\caption{Standard deviation (STD) used in relative uncertainty calculations}
\label{tab:data-std}
\end{table}



\end{appendices}

\FloatBarrier


\bibliography{refs}

\end{document}